**Linear-scaling kernels for protein sequences and small molecules outperform deep learning while providing uncertainty quantitation and improved interpretability**

Jonathan Parkinson[1] and Wei Wang[1,2]
[1]Department of Chemistry and Biochemistry, [2]Department of Cellular and Molecular Medicine, University of California, San Diego, La Jolla, CA 92093-0359

Correspondence to: Wei Wang, email address: wei-wang@ucsd.edu

**ABSTRACT**
Gaussian process (GP) is a Bayesian model which provides several advantages for regression tasks in machine learning such as reliable quantitation of uncertainty and improved interpretability. Their adoption has been precluded by their excessive computational cost and by the difficulty in adapting them for analyzing sequences (e.g. amino acid sequences) and graphs (e.g. small molecules). In this study, we introduce a group of random feature-approximated kernels for sequences and graphs that exhibit linear scaling with both the size of the training set and the size of the sequences or graphs. . We incorporate these new kernels into our new Python library for GP regression, xGPR, and develop an efficient and scalable algorithm for fitting GPs equipped with these kernels to large datasets. We compare the performance of xGPR on 17 different benchmarks with both standard and state of the art deep learning models and find that GP regression achieves highly  competitive accuracy for these tasks while providing with well-calibrated uncertainty quantitation and improved interpretability. Finally, in a simple experiment, we illustrate how xGPR may be used as part of an active learning strategy to engineer a protein with a desired property in an automated way without human intervention.

# INTRODUCTION

Identification of proteins and small molecules with desired properties is a task of crucial importance for the pharmaceutical and chemical industries[1]. In recent years, machine learning assisted approaches have become more popular[2–8]. Neural network/deep learning models are widely used owing to their flexibility, their ability to learn complex relationships from large datasets, and – owing to the development of libraries like PyTorch – ease of implementation[2].

Despite deep learning's success, it suffers from several limitations. First, most deep learning architectures do not quantify their uncertainty on predictions, tend to be "over-confident" when extrapolating and are "black-box", so that it is very hard to determine why the model makes a particular prediction for a specific input[9]. Second, surprisingly, deep learning models have repeatedly been shown to lack robustness to so-called "adversarial attacks", small perturbations to their input. Adding subtle noise not perceptible to the human eye, for example, could reliably cause deep learning models to misclassify a photo of a cat as guacamole, or a lionfish as eggnog[10]. In biology, protein structure prediction models have been shown to exhibit the same vulnerability to adversarial attacks[11]. Finally, deep learning often entails enormous computational cost. Many state of the art models use tens or hundreds of billions of learned parameters. Often a large fraction of their parameters can be removed without damaging performance, suggesting a more efficient approach is possible. Xu et al., for example, pruned *97%* of the parameters of the BERT model in natural language processing and yet achieved equivalent performance[12].

An alternative model architecture is Gaussian process (GP) regression[13]. A GP is a Bayesian model which defines a multivariate normal distribution over possible functions mapping the input x variable to an output y. In regions where there is little or no training data, the model expresses high uncertainty. In regions where there is substantial training data, the model predicts outcomes with more confidence. The type of functions in the distribution is determined by a kernel function that needs to be selected and its hyperparameters "tuned" (as with any other model) for a specific problem. The kernel function measures the similarity of any two x-inputs.

GP models have at least four compelling advantages. First, like deep learning, if equipped with an appropriate kernel, a GP is able to approximate any relationship[13]. Second, GP can calculate the marginal likelihood – the probability of the data averaged over all possible function values – in closed form. Consequently, kernel hyperparameters can be "learned" by maximizing marginal likelihood rather than the likelihood, reducing the risk of overfitting and thus making the model more robust. Third, the model's predictions are generated using the similarity of new data points to those in the training set, where similarity is quantified by the kernel function. Unlike a deep learning model, a GP is not a black box, because one can determine how the model measures similarity between datapoints and generates predictions. As we will show, the same kernel function that the GP model uses to make predictions can also be used to cluster the training data and retrieve the most "similar" datapoints in the training set, where "similarity" is determined by the kernel function we selected.

Fourth, a GP reliably quantifies its uncertainty and assigns high uncertainty to datapoints very dissimilar from its training set. Uncertainty is important for protein engineering and drug design where experimentally evaluating model predictions is expensive. In these scenarios, one would rather use only model predictions that are more likely to be reliable. GP would detect "distribution shifts", where new data points are very different from the training set by assigning high uncertainty to its predictions, while a deep learning model may "fail silently", generating low-accuracy predictions without providing any obvious sign of failure[9,14,15]. Uncertainty can also be used for Bayesian optimization / active learning, where the model assists the practitioner in selecting which datapoints to experimentally evaluate next[16–18]. Several techniques for estimating uncertainty in deep learning have been introduced in the literature[19–22]. As we will show here, however, the uncertainty estimates they provide are much more poorly calibrated than those provided by a GP.

GPs suffer from several limitations. The ability to choose the kernel function can be both a drawback and a benefit, since there are many problems for which we do not have enough prior knowledge to design an appropriate kernel (e.g. image classification). It may be possible in some cases to overcome this drawback by combining deep learning with Gaussian processes, and indeed this possibility has been explored by some practitioners in the literature [23,24]. For example, a deep learning model can be trained in an unsupervised fashion to learn a feature representation which can be used as input to a GP. In this strategy, the GP essentially serves as the last layer of the deep learning model, thereby hopefully garnering some of the benefits of both approaches. We experimentally evaluate one strategy for combining deep learning with a GP in this paper as well.

The main challenge for GP is computational expense. A straightforward implementation scales with dataset size as $O(N^3)$[13]. Indeed, the poor scaling of GPs is often cited as a reason to prefer deep learning. Li et al., for example, when building a model for traffic prediction, argued that it was not possible to use a GP for their problem because "GPs are hard to scale to a large dataset"[25]. A second major challenge is the lack of efficient kernels for sequence and graph data. Many kernels for graphs have been described[26], but typically exhibit quadratic or worse scaling in the size of the graph.

One approach to approximating a GP is the random Fourier feature (RFF) approach of Rahimi and Recht[27]. Briefly, RFF methods approximate the kernel function via a random map corresponding to the kernel of interest, so that each input datapoint is converted into a "random feature" representation; one is then able to approximate the GP using Bayesian linear regression in the random feature space, and thereby achieve linear scaling with training set size.

In this paper, we set out to develop approximate Gaussian process models that can achieve good scalability and performance for sequences and graphs. We begin by introducing efficient methods for fitting random features-approximated Gaussian processes using preconditioned conjugate gradients (PCG) with a new preconditioner. We demonstrate the scalability of the resulting approximate GP to larger datasets, show that it outperforms an alternative method for approximating Gaussian processes on 9 different datasets, and find that it achieves the same test set performance as an exact (non-approximated) GP on a protein engineering task.

Next, we introduce a group of efficient, scalable, random feature-approximated kernels for graphs and sequences. We compare the test set performance of GP regression using these kernels with deep learning methods on 17 different benchmarks. The GP regression models achieve highly competitive performance..

We explore the possibility of combining deep learning with GPs by using embeddings as input to our approximate GP and show that this is frequently beneficial. We demonstrate that uncertainty estimates provided by an approximate GP are significantly more well-calibrated than those provided by a variety of uncertainty estimation techniques for deep learning. In a simple experiment, we show how an approximate GP can be used to "engineer" a protein with improved fitness with no human intervention required. Finally, we illustrate how the approximate Gaussian process can be used to cluster the training set and retrieve the datapoints in the training set most similar to a test point.

## METHODS

### The random features approximation: background

In a typical Gaussian process model (see ref [13] for details), given a training set X with associated y-values, we make predictions for the y-value $f_*$ associated with a new datapoint $x_*$:

$$p(f_* \mid X, y, x_*) = N(u_*, \sigma_*) \tag{1}$$

where $N(u_*, \sigma_*)$ is a Gaussian distribution with mean and variance of

$$u_* = k_*^T (\lambda^2 I + K)^{-1} y \quad \text{and} \quad \sigma_* = k_{**} - k_*^T (\lambda^2 I + K)^{-1} k_* \tag{2}$$

Here $k_*$ is a vector of length *N* for *N* training datapoints formed by evaluating the kernel function $k(x_*, x_i)$ for each $x_i$ in the training set, and *K* is an N x N matrix where element *ij* is formed by evaluating the kernel function for $k(x_i, x_j)$ points in the training set. $\lambda$ is a kernel hyperparameter which denotes the level of expected noise in the data. $k_*$ is the kernel function evaluated on the new datapoint against all training datapoints, and $k_{**}$ is the kernel function evaluated on the new datapoint against itself. When fitting, we do not have to "learn" any parameters, but we *do* have to construct and invert an *N x N* matrix, which is expensive if N is large.

To avoid this cost, we approximate kernel functions using random Fourier features[27]. This well-known approximation specifies a random map for a given kernel such that $k(x_1, x_2) \approx z(x_1)^T z(x_2)$ (for details, see Supporting Information Section S1). The mean and variance are now predicted for new datapoints using:

$$\mu_* = z_*^T [Z^T Z + \lambda^2 I]^{-1} Z^T y = z_*^T w, \qquad \sigma_* = \lambda^2 z_*^T [Z^T Z + \lambda^2 I]^{-1} z_* = \lambda^2 z_*^T V z_* \qquad (3)$$

Where Z is the transformed input matrix, such that row *i* is $z(x_i)$ and $z_*$ is $z(x_*)$ where $x_*$ is a new datapoint, λ is a kernel hyperparameter. For common stationary kernels, the error of the approximation decreases exponentially with an increasing number of random features[27], so that there are diminishing returns (increasing the number of random features from 1,000 to 2,000 achieves a larger relative improvement than going from 2,000 to 4,000 and so forth).

We further replace vanilla random features described above with the structured orthogonal random features (SORF) procedure[28]. Briefly (for details, see Supporting Information Section S2), with the RBF kernel as an example, the following transformation is applied to each input vector:

$$SHD_1 HD_2 HD_3 \qquad (4)$$

Where S is a diagonal matrix with elements drawn from a χ-distribution with *d* degrees of freedom, H is the normalized Hadamard matrix, and $D_1, D_2, D_3$ are diagonal matrices with elements drawn from the Rademacher distribution. The Hadamard matrix multiplication can of course be replaced with a fast Hadamard transform[29], reducing the cost of generating random features from $O(NM^2)$ to $O(NM \log M)$. The resulting model is very lightweight, since we only need to store four diagonal matrices. Generating predictions with a trained model is fast, since we need only perform a series of transforms & diagonal matrix multiplications, then take a dot product. This alteration is not only beneficial for speed but also for performance. Yu et al[28] demonstrate that this modification improves the accuracy of kernel approximation with random features for the RBF kernel. We implement the fast Hadamard transform for both CPU and GPU in our library; see Supporting Information section S3 for benchmarking and other details.

**Faster fitting with a new preconditioner**

To fit the approximate GP model, we must find $[Z^T Z + \lambda^2 I]^{-1} Z^T y = w$. If the number of random features is large (e.g. 5000 - 30000), directly forming and inverting the matrix $[Z^T Z + \lambda^2 I]$ requires expensive matrix multiplications and decompositions. We avoid this problem by instead using the well-known method of conjugate gradients (CG)[31], which achieves subquadratic scaling in the number of random features and iteratively solves $Z^T y = [Z^T Z + \lambda^2 I] w$ for *w*. Briefly (for details, see Supporting Information section S4), rather than computing

$[Z^T Z + \lambda^2 I]^{-1}$, this algorithm requires that on each iteration, we compute the matrix-vector product $[Z^T Z + \lambda^2 I]w$. No matrix multiplications or decompositions are required at any time.

Note that $[Z^T Z + \lambda^2 I]w = (\sum_{i}^{N} z_i^T z_i + \lambda^2 I)w$ for N row vectors in Z; in other words, the matrix vector product can be formed as a sum of products over individual rows. We therefore process the dataset in minibatches. For each minibatch, we generate the random features using the Hadamard transform-based procedure and perform a matrix-vector multiplication with $w$ then add this to the result. Our implementation also allows easy parallelization. On each iteration of conjugate gradients, each worker can be assigned its own copy of the kernel, its own copy of $w$, and its own subset of the data to process. At the end of the iteration, the results from all the workers are summed and used to update $w$ for the next iteration.

Note that unlike with stochastic gradient descent-based fitting methods, minibatch size has no effect on model performance or on the number of iterations required for convergence, because CG evaluates the full gradient on each iteration. Larger minibatches can slightly reduce the time required to fit but do increase memory footprint. Thus the memory footprint is independent of the dataset size, and can be determined by the user by choosing minibatch size.

Since CG is iterative, it is important to minimize the number of iterations to make model fitting as fast as possible. xGPR uses the randomized Nystrom approximation[30] of the matrix $[Z^T Z + \lambda^2 I]$ in equation (1) as a preconditioner (for details, see Supporting Information Section S4). While this is a known method for matrix approximation, it has not previously been used as a preconditioner for Gaussian processes to our knowledge.

We also add a modification to the randomized Nystrom approximation that reduces the time needed to construct the preconditioner (see Supporting Information Section S4 for details and for benchmarks). Importantly, in our implementation the preconditioner is constructed by loading only one minibatch of the data into memory at a time, minimizing our memory footprint. We finally show in detail (see Supporting Information S5 and S6) how to use random features and more specifically preconditioned conjugate gradients for hyperparameter tuning.

In Figure 1, we compare our preconditioned conjugate gradient (PCG) approach with the Scipy library's implementation of L-BFGS across 7 datasets; L-BFGS (limited memory BFGS) is a popular algorithm for minimizing the loss function with an iteratively constructed approximation to the Hessian[34]. The PCG strategy reduces the number of iterations required for convergence and hence the time to fit by as much as *two orders of magnitude* compared with L-BFGS or with non-preconditioned CG, enabling us to achieve a tight fit in as few as 10-20 iterations.

In Figure 2 we further compare our PCG approach with two stochastic gradient descent methods, stochastic variance reduced gradient descent (SVRG) and AMSGrad, an improved variant of the Adam algorithm which is popular for training neural networks. (In initial experiments, Adam performed worse than AMSGrad and is thus not further evaluated here). We illustrate how the residual error decreases with the number of epochs (full passes over the dataset). For PCG, each iteration is equivalent to one epoch, while for stochastic gradient descent, there are many iterations per epoch. As illustrated, PCG is the clear winner, converging faster than stochastic gradient descent by orders of magnitude and thereby greatly reducing the number of epochs required to fit. Unlike stochastic gradient descent, CG does not require tuning of a learning rate to achieve good results. Our PCG approach coupled with our random feature generation method enables us to fit a > 400,000 datapoint dataset in less than 7 minutes *including* all hyperparameter tuning. If using a minibatch size of 2,000 datapoints, the process requires less than 1.5 GB of RAM at all times.

Note that fitting half a million datapoints using an exact Gaussian process would require constructing a 400,000 by 400,000 kernel matrix, which would require 1.3 terabytes of memory, and the entire kernel matrix would be required to fit the model. xGPR is thus much more scalable than an exact Gaussian process. Indeed, as we will see below under "Comparison with stochastic variational inference and exact Gaussian processes", when using an exact Gaussian process as implemented in the GPyTorch library, we immediately encountered an out of memory error when trying to fit a dataset with merely 21,000 training datapoints, while xGPR was able to fit half a million datapoints without any issues.

# **RESULTS**

**Comparison with stochastic variational inference and exact Gaussian processes**

Random features is not the only method for approximating a Gaussian process; one alternative is stochastic variational inference (SVI), which uses a set of "inducing points" much smaller than the number of training datapoints to approximate the GP[31]. In the Supporting Information section S8, we fit both xGPR and a GP approximated using stochastic variational inference (as implemented in GPyTorch[32]) to 9 different datasets ranging in size from 400 to half a million datapoints. We use the same kernel in all cases to ensure a fair comparison (See Supporting Information section S7 for details on these and all other datasets used for benchmarking in this paper).

These experiments were performed on a GTX1070 GPU with 8 GB of RAM. Two of the datasets are sufficiently small (< 5,000 datapoints) that we can fit an exact Gaussian process, and so we fit an exact GP for those two datasets as well. We tried fitting an exact GP to a third dataset with about 21,000 datapoints, but the GPyTorch library immediately threw an out of memory error, demonstrating that an exact GP is already expensive for datasets of fairly modest size.

We find that xGPR achieves the same or better performance as SVI *in every single case*; there is no case where SVI outperforms xGPR. For the two datasets where we can compare xGPR with an exact GP, xGPR achieves very similar performance, while SVI is similar in one case and in the other significantly worse.

Finally, we compare xGPR with an exact Gaussian process on an active learning / Bayesian optimization task from the literature. Sarkisyan et al. 2016[33] generated data for 52,910 mutants of the green fluorescent protein (GFP), measuring the fluorescence of each mutant. Hie et al.[23] fitted an exact Gaussian process with an RBF kernel to this dataset, using 1,115 sequences with only one mutation as their training set and representing the input datapoints using learned embeddings. In Bayesian optimization, an acquisition function is used to select new datapoints for experimental evaluation. Hie et al. use the upper confidence bound (UCB), which is the predicted mean value of the GP, plus a confidence interval. UCB prioritizes both datapoints where the model predicts the mutant sequence will have a high fitness value and datapoints where the model has high uncertainty, thus achieving a balance between exploitation of existing knowledge and exploration of sequence space. They show that the UCB values for the test set are reasonably well-correlated with the actual measured fluorescence values.

We fit xGPR to the same dataset, using the same training set, kernel and representations. Hie et al. found the UCB acqusition function exhibited a Spearman's-r correlation coefficient of 0.78 with measured fluorescence in the test set. Surprisingly, the result for xGPR is slightly better at 0.8. Next, we pick the 50 mutants from the test set with the largest UCB values (to simulate the effect of experimentally assessing the model predictions). The average log fluorescence of the top fifty for xGPR is about the same as the result from Hie et al (both are about 3.75). These results suggest that when using the same kernel, xGPR compares well with an exact Gaussian process while achieving greatly improved scalability.

### *A first new set of efficient kernels for sequences that cannot be aligned*

Up to this point, we have only considered kernels for fixed-length inputs. In protein engineering problems, it is often possible to form a multiple sequence alignment from all of the input sequences. Alternatively, if a learned embedding generated by a large language model is used to represent the input, then the learned embedding can be averaged over the sequence to generate a fixed-length input.. In many other cases, however, the sequences are of different lengths and cannot easily be aligned – if the input sequences are from different families, for example. We now introduce two types of random features approximated kernels that can efficiently handle this problem. We refer to the first type as FHT-Conv-1d and the second as Fast-Conv-1d.

The first type of kernel can be constructed as follows. Consider extracting all k-mers of a given length, e.g. 9, from two sequences, and calculating the similarity of each possible pairing of k-mers from the two sequences using any common fixed-vector kernel of interest. There are a range of different kernels we could use for such a comparison, including the RBF, Matern, neural network and arc-cosine kernels, all of which can be approximated using random features. (The "convolutional kitchen sinks" kernel of Morrow et al.[34] is a special instance of this type of kernel.) If we use the this case the comparison between the two sequences is as follows:

$$\sum_i^{d_1-n} \sum_j^{d_2-n} k(x_{i:i+n}, y_{j:j+n}) \tag{5}$$

Where *k* is the fixed-vector kernel we have selected to measure the similarity of any two subsequences. As written, this kernel scales as the square of the sequence length. . It is therefore impractical if used in an exact Gaussian process or in stochastic variational inference, because in those cases it will need to be implemented as written, resulting in a very large number of pairwise k-mer comparisons. If comparing two protein sequences of length 250 with subsequence length 9, for example, we would need to extract all 242 k-mers from both sequences then perform *over 29,000 pairwise k-mer comparisons*.

We now show how to use the random features approximation to implement this kernel in a way that scales linearly with sequence length. Using the random features approximation, this kernel becomes:

$$\sum_i^{d_1-n} \sum_j^{d_2-n} k(x_{i:i+n}, y_{j:j+n}) \approx \sum_i^{d_1-n} \sum_j^{d_2-n} z_{i:i+n}^T z_{j:j+n} \tag{6}$$

Where the two z-vectors are the random features generated for each pair of k-mers in sequences *x* and *y*. The last expression shown above is equivalent to:

$$\left(\sum_{i}^{d_1-n} z_{i:i+n}\right)\left(\sum_{j}^{d_2-n} z_{j:j+n}\right) \quad (7)$$

We can therefore generate random features for each subsequence of length *k* in the two input sequences and sum them, resulting in a single vector representing sequence *x* and sequence *y*, then use these representations as inputs for Bayesian linear regression. The computational cost of this approach exhibits linear scaling with both dataset size and sequence length. This can be considered a convolution kernel since generating the random features for the k-mers in each input sequence can be accomplished through convolution with filters populated with random weights drawn from the distribution appropriate for the kernel we want to approximate. This approximated kernel still has however limited practicality since convolution with thousands of filters is slow.

We now introduce a second innovation that greatly accelerates the calculation of random features for these kernels. To understand our implementation, first note that in deep learning libraries convolution is often implemented using matrix multiplication as follows. Consider an input *K x P* matrix with *P* sequence elements (e.g. *P* amino acids) and *K* features per element (or amino acid), and let us assume our convolution filter is of width *L*. We can use the input matrix to populate an *LK x (P - L + 1)* matrix where each column *i* is the concatenated features associated with subsequence *i*. The *m* convolution filters meanwhile can be stacked to form an *m x LK* matrix; multiplying this matrix against the *LK x (P - L + 1)* input matrix is equivalent to performing 1d convolution with *m* filters of length *L*.

In our implementation, we replace the *m x LK* matrix of convolution filters with the SORF operation (equation 4). We thereby replace matrix multiplications with fast diagonal matrix multiplications and fast transforms, analogous to replacing a discrete Fourier transform with FFT. For simplicity, let the number of random features desired, *m*, be the same as *LK*. Then the implementation of this kernel with vanilla random features scales as $O(N(P - L + 1)m^2)$. The proposed modification by contrast scales as $O(N(P - L + 1)m \log(m))$.

Depending on the fixed-vector kernel we use to measure distances between k-mers, we can choose how we encode the amino acids in the two input sequences to this class of kernels so that distance between k-mers is measured in a particular way. If we one-hot encode amino acids, for example, the distance between any two k-mers is twice the Hamming distance. Alternatively, we can represent each amino acid using a learned representation, e.g. from a large language model.. We explore these possibilities below under "Evaluating convolution kernels for sequence data". We refer to this first type of kernel as FHT-Conv-1d.

**A second efficient convolution kernel for sequences which cannot be aligned**
We now introduce a second efficient group of convolution kernels which mimic a three-layer neural network, which we call Fast-Conv-1d. We make use of the following simple lemma (see the Supporting Information Section S9 for a proof):

*Lemma 1.1.* Let $K: \mathbb{R}^D \times \mathbb{R}^D \rightarrow \mathbb{R}$ be a positive definite kernel on $\mathbb{R}^D$, and let $f: X \rightarrow \mathbb{R}^D$ be any mapping from $X$ to $\mathbb{R}^D$, where $X$ is some non-empty set. Then $K(f(u), f(w))$ is a positive definite kernel for all $u, w \in X$.

Consider a three-layer neural network with random weights where the first layer is a convolutional layer with ReLU activation followed by global max pooling and the second layer is fully-connected with a selected activation function. This is analogous to a fixed vector kernel (e.g. RBF or Linear + RBF) whose input is a random feature convolution with ReLU activation and global max pooling (Figure 3). The fixed vector kernel is of course approximated using a fast Hadamard transform-based random features procedure. This combination kernel is positive definite following lemma 1.1. This kernel has the property that the convolutions can all be performed exactly *once* before training and the results saved on disk. We can use the fast Hadamard transform-based procedure outlined above to perform the random feature convolutions in an efficient way.

To understand how this type of kernel quantifies the similarity of sequences, consider performing a convolution of a one-hot encoded sequence with a width-9 filter containing weights drawn from a normal distribution. For each subsequence of length 9, the convolution will be large and positive *if* the filter happens to contain large positive values at the same positions that are one-hot in the subsequence. Since we perform ReLU activation and max-pooling across the sequence, for each convolution filter, the resulting feature is the largest positive value encountered – in other words, the best "match" for that filter found in the sequence. Thus, this kernel essentially "profiles" each sequence input to quantify the best match in the input for each filter in a set of random filters, then compares the profiles of any two input sequences using a specified fixed vector kernel (e.g. RBF or Linear + RBF; we use RBF in all remaining experiments for simplicity). As before, we can encode sequences using learned representations, one-hot encoding, or some other scheme as desired.

**Evaluating xGPR and convolution kernels for sequence data**

We evaluate xGPR and our FHT-Conv-1d and Fast-Conv-1d kernels by fitting protein engineering benchmark data from Rao et al. and Dallago et al.[35,36] (the TAPE benchmarks and FLIP benchmarks respectively). These datasets involve situations where we have been provided with a pre-constructed training set and test set consisting of protein sequences, and the goal is to predict a real-valued outcome for each sequence. We describe these datasets briefly here (for more details on these and all other datasets used in this paper, see Supporting Information Section S7).

The AAV dataset[37] involves mutations including insertions and deletions to a 28-amino acid window of an adeno-associated virus capsid; some of these mutations were randomly generated while others were "designed" (selected intentionally). Dallago et al. split the data up in various ways and refer to these various partitions as "splits". For example, they train a model on the randomly mutated proteins and test on the designed ones ("mutant vs designed split"), or train on the designed proteins and test on the mutated ones ("designed vs mutant split"), or train on proteins with seven mutations only and test on the rest ("seven vs rest split"), etc. The goal is to predict the fitness of the mutant, where higher is better.

The thermostability dataset[38] measures the thermal stability (higher is better) of 48,000 proteins from 13 species using a mass spectrometry approach; the resulting proteins are highly diverse. We use two splits here, "mixed" (including all species) or "human" (a smaller subset including human data only). The GB1 dataset[39] measures the fitness of mutants of the binding domain of protein G from a species of bacteria, with mutations at just four sites in the protein. The dataset includes almost complete coverage, i.e. most of the possible four-site mutants were generated and tested. The "three vs rest" split uses mutants with three mutations only as the training set and all others as test, while "two vs rest" uses mutants with two mutations only as training.

The fluorescence dataset from TAPE involves mutations to the green fluorescent protein; here the goal is to predict log fluorescence for test set proteins based on a training set. This dataset is the same as the one used by Hie et al. discussed above under "Comparison with stochastic variational inference and exact Gaussian processes", although the TAPE benchmark splits the data into training and test differently. Finally, the stability dataset from TAPE[40] measures the stability of a diverse set of proteins in a protease susceptibility assay, where a larger value indicates better stability.

We fit in triplicate using different random seeds; the resulting error bar measures variability across different random seeds. Since our focus here is on Gaussian process regression, we use only benchmarks which involve regression tasks.

In Table 1, we compare the FHTConv-1d kernel and FastConv-1d kernels trained on one-hot encoded data with convolutional neural networks trained on one-hot encoded data by Rao et al and Dallago et al. In Table 2, we use learned embeddings – representations of the input sequence generated by the FAIR ESM-1v pretrained language model for protein sequences – to represent the input sequences, and we compare our results with those of fine-tuned language models reported by Rao et al and Dallago et al. We report the *best* result from any fine-tuned large language model described by Rao et al. and Dallgo et al. for each benchmark.

For the learned embeddings, there are at least two ways we can use them as input to a Gaussian process, and we evaluate both. First, we can average the representation for each amino acid across all amino acids in the sequence as done by Hie et al. among others then use an RBF kernel. Alternatively we can arrange the learned embeddings of size D for each sequence of length M as an M x D array and use our FHTConv-1d or FastConv-1d kernels to perform convolution across the sequence. It is somewhat easier to train using the FastConv-1d kernel since the convolutions can all be performed once at the start of training, thus eliminating the need to save the learned embeddings to disk (which can be quite large). We therefore use FastConv-1d here.

We note here three interesting findings. First, xGPR when trained on one-hot encoded input with a modest number of random features (< 10,000) outperforms a CNN on 8 out of 11 benchmarks and ties on one more. When trained on learned embeddings, the approximate GP outperforms the fine-tuned language models on 6 benchmarks and ties on one more. Note that the pretrained language models required up to 50 GPU-days training time to fit, while the FastConv-1d kernel takes less than 6 minutes per fit to fit every dataset shown here (including hyperparameter tuning), with a maximum memory footprint of 1.7 GB when using a minibatch size of 2000 datapoints. (The FHTConv-1d kernel is slower but always takes less than 15 minutes including hyperparameter tuning). Using learned embeddings generally improves the performance of the GP, *especially* on diverse datasets with sequences from many different protein families (e.g. thermostability), but not always; on several benchmarks the GP performs better with one-hot encoding.

Second, using a larger number of random features in all cases slightly improves the performance of the Gaussian process; this is unsurprising since it improves the accuracy of the kernel approximation. This feature provides the user with a simple way to improve performance if desired. Third, we note that neither the FHT-Conv-1d kernel nor the Fast-Conv-1d kernel is *consistently* superior to the other; rather, each outperforms the other on some of the datasets.

.

**Uncertainty calibration**

A major motivation for using a GP is to quantify uncertainty. Various methods for uncertainty estimation for deep learning have been reported in the literature. We now show that xGPR provides better-calibrated estimates of uncertainty than these alternatives.

For an uncertainty estimate to be useful, we want uncertainty to increase as we move far away from the training set – this is true for a GP with a stationary kernel by construction[13]. Additionally, we desire an uncertainty estimate that is *well-calibrated*, in other words, the uncertainty reflects how often a predicted outcome will actually happen. If a weather forecast claims there is a 10% chance of rain, for example, there should be rain on ten out of a hundred days where it makes this prediction. Likewise, for a regression model, we would expect that roughly 10% of the test datapoints will lie within a 10% confidence interval on its predictions (or a "credible interval" in Bayesian inference), roughly 20% of datapoints will lie within a 20% confidence interval, 90% within a 90% confidence interval, and so on.

Calibration can be quantified as area under the calibration error curve (AUCE), introduced by Gustafsson et al[41]. For 100 values in the range from 0 to 1 (0.01, 0.02, etc.) we determine what fraction of the test set predictions lie within the corresponding confidence interval (1%, 2% etc.) We subtract the observed fraction within the interval from the expected fraction and take the absolute value – for example, if the confidence interval is 10% and the observed fraction within that interval is 90%, the difference would be $|0.9 - 1|$. Finally, we calculate the area under this curve; the larger the area, the more poorly calibrated the uncertainty figures provided by the model. A perfectly calibrated model achieves a score of 0.

Greenman et al.[42] use five different methods for uncertainty estimation (dropout uncertainty[20], last-layer stochastic variational inference[43], model ensembling, deep evidential regression[21] and mean-variance estimation) in combination with deep learning models on five of the benchmarks we have used in this paper. They use both one-hot and ESM embeddings as input to their models and calculate AUCE as described. In Table 3, we compare the AUCE for their uncertainty estimation methods with the AUCE for xGPR for the same datasets using the same encodings. To give the deep learning models the benefit of the doubt, we present the *best* AUCE reported for each uncertainty estimation technique reported by Greenman et al. for each dataset, and we use the xGPR model from Tables 1 and 2 that displayed the best performance for that dataset (as assessed by correlation between predicted and measured fitness).

Where uncertainty calibration is concerned, on all five benchmarks, we see that xGPR performs substantially better, with an AUCE from threefold to tenfold better than the best deep learning uncertainty estimation method. The AUCE for xGPR is < 0.15 in every case, while the deep learning methods often (and for some methods always) achieve AUCE scores of 0.3 or above.

**Using uncertainty for protein engineering**

Uncertainty can of course be used to determine whether a prediction generated by a model should be trusted; if the uncertainty is high compared to uncertainty for the training set, we may consider a prediction less likely to be reliable. Alternatively, we can use the uncertainty estimates generated by xGPR as part of a Bayesian optimization strategy. In this framework, we first fit the model to the available data. We then choose a new batch of sequences to experimentally evaluate using an acquisition function that takes into account both the model predictions and our uncertainty, thereby achieving a balance between exploration and exploitation. The upper confidence bound (UCB), for example, adds the upper half of the confidence interval to the model prediction for each test point. After experimentally evaluating the selected points, we re-fit the model and repeat this process until a desired protein function is achieved. For an overview of other common acquisition functions, see Frazier 2018[44].

In this experiment, we use the GB1 dataset discussed above under "Evaluating convolution kernels for sequence data", in which mutations are introduced to four sites in the binding region of protein G. This dataset is convenient for an experiment of this kind because close to 150,000 of the 160,000 possible mutants have been experimentally evaluated. We normalize the fitness of all mutants in the dataset so that it lies between 0 (worst possible) and 1 (best possible). Note that > 99.95% of the sequences have a fitness < 0.6; only one has a fitness of 1. We randomly select 384 variants and train an xGPR model with an RBF kernel, then generate the UCB acquisition function score for all of the remaining mutants and use it to select a batch of 96 to "experimentally" evaluate. We then add these 96 to the training set, refit the model and repeat the process for up to five iterations. At each iteration, we keep track of the best fitness achieved so far. Finally, we repeat this experiment 50 times with different random seeds.

By the fifth iteration, all repeats achieved a best fitness of 0.6 or above. 40 of the 50 repeats had found one of the four best sequences, and 30 of the repeats had achieved a score of 1 – i.e. had found the best sequence in the dataset – despite experimentally testing < 1000 sequences. Half of the repeats were able to find the best sequence (fitness 1) in four iterations or less. These results suggest that if combined with a high-throughput experimental system, this approach using xGPR might be able to "discover" proteins with desirable properties with no human intervention required.

**Efficient, numbering-invariant kernels for small molecules and other graphs**
We introduced two groups of linear-scaling kernels for sequences under "A first new set of efficient kernels for sequences that cannot be aligned" above. These kernels can easily be extended to graphs, if we encode each node in the graph as a feature vector containing some information about that node, its edges and adjacent neighbors. For the Fast-Conv-1d kernel, we can apply the random feature convolution and max-pooling operation across all feature vectors associated with the graph. For the FHT-Conv-1d kernel, we can merely use nodes and their associated features as inputs rather than k-mers, and perform a width-1 convolution across the nodes of the graph; we refer to this as a Graph-Conv-1d kernel.

We implement several variants of the Graph-Conv-1d kernel in xGPR. We focus here on the simplest variant, Graph-RBF, which approximates a pairwise comparison across the nodes of two graphs using an RBF kernel. To evaluate the performance of Graph-RBF, we consider the prediction of energetic and thermodynamic properties of small molecules. We benchmark on the QM9 dataset of Ramakrishnan et al.[45], which contains 134,000 small molecules whose geometry has been optimized using density functional theory (DFT). The kernels we describe here seem especially likely to be a good fit for energetic / thermodynamic properties based on prior knowledge, since these properties can in principle be decomposed as a sum of per-atom contributions.

We use two different modeling approaches. In approach (1), we one-hot encode each atom *i*, then concatenate to this a one-hot encoding of each other atom *j* in the molecule divided by the distance of atom *j* to atom *i* to the 6th power (to emulate the London matrix of Huang et al[46]); the one-hot encodings are sorted by distance. Because of the 6th power weighting, out past four or five angstroms the features associated with more distant atoms become negligibly small and can be disregarded. We then use this representation as input to a Graph-RBF kernel. This simple representation is easy to construct but is not very informative (it discards all information about bond angles) and thus unlikely to achieve good performance.

Alternatively, in approach (2), we represent each atom using the SOAP features of Bartok et al.[47] In this scheme, the atomic density surrounding each atom is expanded in an orthonormal basis set, and the power spectrum of these coefficients is used as a representation. We use the distance weighting scheme described by Willatt et al.[48] with only small modifications to their selected hyperparameters. For more details on the hyperparameters used and on hyperparameter tuning see Supporting Information section S10.

Remarkably, approach (1), despite using uninformative features, achieves chemical accuracy of about 1 kcal/mol. (2) is much better. For internal energy, enthalpy and Gibbs free energy, xGPR can achieve an MAE < 0.2 kcal/mol, and for one of the six properties we achieve a new state of the art. Note that 1 kcal/mol is typically considered to be "chemical accuracy", i.e. the accuracy required to make correct predictions for the behavior of chemical systems, so that both xGPR and several of the other models are well above the level of accuracy that is actually needed. Indeed, the density functional theory (DFT) calculations whose outcome we are trying to predict are often only accurate to 2-3 kcal/mol. Consequently, a difference of 0.01 - 0.05 kcal/mol between models for predicting DFT-calculated energies is likely equivalent for practical purposes.

**Retrieving similar datapoints from the training set**
We can measure feature importance using the SHAP or LIME techniques for a Gaussian process just as for any other ML model. Unlike for other ML models, we can also for a small dataset additionally construct the kernel matrix and use it to determine exactly how much each datapoint in the training dataset contributed to a given prediction. This kind of decomposition is not possible with a deep learning model.

Constructing the kernel matrix explicitly however is too expensive if the dataset is larger than 5,000 datapoints. Also, for large datasets assigning a precise contribution to each point in the training set may be more information than is required or helpful. Often, practitioners may merely want to know what are the *most similar* datapoints in the training set – the ones which contributed the most to a prediction. We can determine this as follows. Let $z_*$ be the random features representation of a test datapoint, and let $z_i$ be the random features representation of training datapoint; then the kernel function for the corresponding test and training datapoints $x_*$ and $x_i$ is approximately $z_*^T z_i$ (see Supporting Information section S1).

In practice, it may be computationally burdensome to take the dot product of a test datapoint representation with *all* of the training data every time we want to retrieve the most similar training datapoints. We can resolve this problem by simply clustering the training data. The same kernel used to fit a Gaussian process regression model can be used to cluster it through random features-approximated kernel k-means clustering. Note that in the same way that random features approximates a Gaussian process as linear regression, clustering the random features-represented training set with k-means approximates applying kernel k-means to the original dataset[49]. Once the dataset has been clustered, if we want to retrieve the most similar training points, we no longer need to measure the similarity of a test point against every training data point, we only need to measure similarity against the datapoints in the nearest cluster.

Tools to perform these functions are included in the xGPR library. We illustrate this procedure for the QM9 dataset, with additional details in Supporting Information section S11. We first fit the QM9 dataset with energy at 298 K as the target using the Graph-RBF kernel with atoms encoded using the one-hot encoding approach (1), then k-means cluster the training set. The resulting elbow plot used to select the number of clusters appears in Supporting Info Figure S6. The elbow plot suggests the training dataset can be broken into five clusters. We next randomly select five molecules from the test set. For each of them, we use the procedure described in the last paragraph to find the fifty most similar molecules from the training set, where similarity is measured by the kernel function. The structures of the query test molecules and the most similar training molecules (as quantified by the kernel using one-hot encoded input) appear in the Supporting Information section S12.

## **DISCUSSION**

Gaussian processes represent a principled Bayesian approach to machine learning that provides straightforward quantitation of uncertainty. Their most important disadvantages are their unacceptable scaling and the lack of efficient kernels for sequences and graphs. We develop efficient approaches for fitting GP models with random features-approximated kernels that achieve linear scaling with training set size and linear scaling with molecule size or sequence length. We implement these tools in a publicly available Python library, xGPR. We demonstrate this library achieves competitive performance on a variety of protein and small molecule property prediction tasks.

As we have shown, a GP equipped with these kernels generates well-calibrated uncertainty estimates on its predictions, unlike any of the deep learning models we have used as comparators. This quality is of considerable importance for protein engineering and drug discovery; since evaluating a prediction is often very expensive, we would prefer if possible to use high-confidence predictions.

Additionally, a GP equipped with these kernels is substantially more interpretable than the comparator deep learning models. For protein sequences and small molecules alike, we are able for each kernel to explain in detail how the model determines the similarity of any two datapoints, which in turn determines how it makes predictions. This enables us to understand the model's limitations and determine how it might be improved. It is not possible to generate a detailed explanation of this kind for any of the deep learning architectures that are currently popular for sequence or graph data; they are all "black box". Further, as we have illustrated for small molecules, we can use the kernel from the trained GP regression model to cluster the input data or perform kPCA, and this clustering may provide additional insights about the distribution of the input data. This feature would be particularly useful for drug design to catalog small molecules into subgroups.

This combination of competitive accuracy, uncertainty quantitation and improved interpretability suggests the xGPR library may potentially be useful for a range of protein and small molecule property prediction tasks.

The kernels we have discussed in this paper all accept a single protein or small molecule as input. For some tasks (e.g. protein-small molecule interaction), it is desirable to use a "pair" kernel that accepts two inputs (a small molecule and a protein sequence for example) and compares them to other pairs. The kernels discussed here can easily be modified to generate several "pair" kernels that may be appropriate for some problems. For example, the first layer of the FastConv-1d kernel we have introduced here can be used to generate a representation of both the small molecule and the protein; these representations are then concatenated and used as input to an RBF kernel. Alternatively, a GraphRBF kernel could be used to generate random features for the small molecule while an FHTConv-1d or FastConv-1d kernel is used to generate random features for the protein; the random features are then concatenated into a single vector to generate the prediction. This last arrangement corresponds to using the sum of the two kernels. It is likely possible to design other efficient kernels specific to pair comparisons of this kind; we defer experimental evaluation of these and other possibilities to future work.

As discussed in the introduction, GPs also suffer from some limitations. We have shown how to design highly scalable GPs for sequences and small molecules, but it nonetheless remains true that in order for a GP to work well, the selected kernel must be appropriate for the problem, which means the practitioner must have prior knowledge of what similarity measures are appropriate. For many problems (e.g. image classification) no such theory exists. Deep learning may be more appropriate for such problems. We have shown that combining GPs with learned embeddings from deep learning can be beneficial for performance, but many questions remain regarding this approach as well. Note for example that while xGPR generally performed better when using an embedding as input in place of one-hot encoding, in several cases the reverse was observed. This is also true for deep learning models; Dallago et al. found that language models sometimes outperformed (and were sometimes outperformed by) a CNN trained on one-hot encoded data. It is not clear at this time what properties of a sequence are captured by a large language model, what properties of the resulting embedding make it most suitable for use with a given kernel, and which learned embeddings are most appropriate for use in a specific task. These and similar questions seem likely to be important for future work.

**Code Availability**
All code used for and needed to reproduce the experiments in this study is available at https://github.com/Wang-lab-UCSD/benchmarking_xGPR . The xGPR library code is available at https://github.com/jlparki/xGPR . The xGPR documentation is available at https://xgpr.readthedocs.io/en/latest/ .

**Data Availability**

All data used for the experiments in this study is publicly available data derived from datasets provided by other authors. For a description of each benchmark dataset and a link at which it can be downloaded, refer to Supporting Information table S7.

**Competing Interests**
The authors have no competing interests to declare.

**Acknowledgements**
The authors would like to express their gratitude to Mina Yao and Young Su Ko for their assistance in testing the xGPR library.

**Table 1. Spearman's-r correlation coefficient of predictions vs ground truth values for convolutional neural networks and xGPR trained on one-hot encoded data**

| Dataset | Split[a] | # training / test datapoints | CNN[d] | FHT-Conv-1d kernel[b] | | Fast-Conv-1d kernel[b] | |
|---|---|---|---|---|---|---|---|
| | | | | 8,192 random features | 16,384 random features | 8,192 random features | 16,384 random features |
| Stability | Stability | 53,614 / 12,581 | 0.51 | 0.61 +/- 0.01 | **0.66 +/- 0.2** | 0.619 +/- 0.01 | 0.611 +/- 0.007 |
| Thermostability[c] | Mixed | 24,817 / 3,134 | 0.34 | | | 0.352 +/- 0.003 | **0.366 +/- 0.003** |
| Thermostability[c] | Human only | 8,148 / 1,945 | 0.5 | | | 0.553 +/- 0.0002 | **0.564 +/- 0.002** |
| Alpha adenovirus (AAV) | Designed vs mutant | 201,426 / 82,583 | 0.75 | 0.764 +/- 0.002 | 0.775 +/- 0.0008 | 0.761 +/- 0.002 | **0.776 +/- 0.0008** |
| Alpha adenovirus (AAV) | Mutant vs designed | 82,583 / 201,426 | 0.71 | 0.733 +/- 0.004 | 0.749 +/- 0.0004 | 0.747 +/- 0.004 | **0.757 +/- 0.002** |
| Alpha adenovirus (AAV) | 7 mutations vs many | 70,002 / 12,581 | **0.74** | 0.712 +/- 0.001 | 0.725 +/- 0.0005 | 0.686 +/- 0.001 | 0.700 +/- 0.002 |
| Alpha adenovirus (AAV) | 2 mutations vs many | 31,807 / 50,776 | **0.74** | 0.569 +/- 0.003 | 0.59 +/- 0.01 | 0.658 +/- 0.004 | 0.674 +/- 0.001 |
| Alpha adenovirus (AAV) | 1 mutation vs many | 1,170 / 81,413 | 0.48 | 0.573 +/- 0.001 | 0.560 +/- 0.002 | 0.781 +/- 0.003 | **0.792 +/- 0.003** |
| Fluorescence | Fluorescence | 21,446 / 27,217 | 0.68 | 0.678 +/- 0.0009 | 0.680 +/- 0.0004 | 0.669 +/- 0.0002 | 0.673 +/- 0.0001 |
| GB1[e] | Three vs many | 2,968 / 5,765 | 0.83 | 0.83 +/- 0.002 | **0.84 +/- 0.002** | | |
| GB1[e] | Two vs many | 427 / 8,306 | 0.32 | 0.68 +/- 0.01 | **0.68 +/- 0.01** | | |

[a.] In many cases, there are multiple "splits" for the same dataset; for the AAV dataset, for example, the "designed vs mutant" split uses designed sequences as a training set and mutants as test, while the "mutant vs designed" split is the reverse. [b.] All convolution kernels use convolution filters of width 9 as a simple default. The best performer for each benchmark is highlighted in bold. For all the models, we repeat the predictions using three different seeds for the random number generator, then report the mean and standard error of the mean. For the scores shown here, higher is better. [C.] For the thermostability dataset, some of the sequences are quite long (thousands of amino acids). The Fast-Conv-1d kernel is much faster for these kinds of sequences since it only needs to perform convolutions once during hyperparameter tuning, so we consider this kernel only here. [D.] All of the results for the one-hot encoded CNN are either from Dallago et al. or Rao et al. [e.] For the GB1 dataset only, the input sequence is four amino acids in length, so for this dataset only we use an RBF kernel instead (equivalent to a convolution kernel of width 4).

**Table 2. Spearman's-r correlation coefficient of predictions vs ground truth values for fine-tuned large language models and for xGPR trained on ESM-1v embeddings**

| Dataset | Split[a] | # training / test datapoints | Pretrained language model, best reported result[d] | RBF kernel, embeddings averaged over sequence[b] | | Fast-Conv-1d kernel trained on per amino acid embeddings[b] | |
|---|---|---|---|---|---|---|---|
| | | | | 8,192 random features | 16,384 random features | 8,192 random features | 16,384 random features |
| Stability | Stability | 53,614 / 12,581 | 0.73 | 0.71 +/- 0.02 | 0.732 +/- 0.008 | 0.67 +/- 0.01 | 0.70 +/ 0.01 |
| Thermostability | Mixed | 24,817 / 3,134 | **0.68** | 0.64 +/- 0.0002 | 0.65 +/- 0.002 | 0.577 +/- 0.004 | 0.587 +/- 0.002 |
| Thermostability | Human only | 8,148 / 1,945 | **0.77** | 0.72 +/- 0.001 | 0.725 +/- 0.001 | 0.706 +/- 0.001 | 0.714 +/- 0.0005 |
| Alpha adenovirus (AAV) | Designed vs mutant | 201,426 / 82,583 | 0.71 | 0.776 +/- 0.001 | 0.784 +/- 0.001 | 0.793 +-/ 0.001 | **0.805 +/- 0.001** |
| Alpha adenovirus (AAV) | Mutant vs designed | 82,583 / 201,426 | 0.79 | 0.706 +/- 0.003 | 0.726 +/- 0.004 | 0.788 +/- 0.004 | **0.798 +/- 0.002** |
| Alpha adenovirus (AAV) | 7 mutations vs many | 70,002 / 12,581 | **0.7** | 0.521 +/- 0.03 | 0.531 +/- 0.01 | 0.632 +/- 0.006 | 0.666 +/- 0.0004 |
| Alpha adenovirus (AAV) | 2 mutations vs many | 31,807 / 50,776 | 0.7 | 0.55 +/- 0.01 | 0.58 +/- 0.03 | 0.742 +/- 0.005 | **0.758 +/- 0.002** |
| Alpha adenovirus (AAV) | 1 mutation vs many | 1,170 / 81,413 | 0.44 | 0.571 +/- 0.009 | 0.573 +/- 0.005 | 0.786 +/- 0.003 | **0.801 +/- 0.006** |
| Fluorescence | Fluorescence | 21,446 / 27,217 | **0.68** | 0.623 +/- 0.002 | 0.634 +/- 0.002 | 0.6606 +/- 0.0002 | 0.667 +/- 0.001 |
| GB1[e] | Three vs many | 2,968 / 5,765 | 0.82 | 0.83 +/- 0.002 | **0.83 +/- 0.002** | | |
| GB1[e] | Two vs many | 427 / 8,306 | 0.55 | 0.71 +/- 0.01 | **0.71 +/- 0.02** | | |

[a.] In many cases, there are multiple "splits" for the same dataset; for the AAV dataset, for example, the "designed vs mutant" split uses designed sequences as a training set and mutants as test, while the "mutant vs designed" split is the reverse. [b.] All convolution kernels use convolution filters of width 9 as a simple default. The best performer for each benchmark is highlighted in bold. For all the models, we repeat the predictions using three different seeds for the random number generator, then report the mean and standard error of the mean. For the scores shown here, higher is better. [D.] All of the results reported for the pretrained language model and one-hot encoded CNN are either from Dallago et al. or Rao et al. These authors report results for multiple pretrained language models; we report the best result they obtained in each case.. [e.] For the GB1 dataset only, the input sequence is four amino acids in length, so for this dataset only we do not use the FastConv-1d kernel and do not average over the sequence.

**Table 3. Area under the calibration error curve (AUCE) for the best xGPR and for four deep learning uncertainty estimation methods using the same input representation; smaller is better**

| Dataset | Split[a] | # training / test datapoints | AUCE, best xGPR model | AUCE, CNNg with dropout for uncertainty | AUCE, CNN with ensemble for uncertainty | AUCE, CNN with evidential uncertainty | AUCE, CNN with mean-variance estimation for uncertainty | AUCE, CNN with last layer SVI for uncertainty |
|---|---|---|---|---|---|---|---|---|
| Thermostability | Mixed | 24,817 / 3,134 | **0.014** | 0.263 | 0.306 | 0.383 | 0.190 | 0.387 |
| Alpha adenovirus (AAV) | Mutant vs designed | 82,583 / 201,426 | **0.108** | 0.336 | 0.393 | 0.373 | 0.348 | 0.438 |
| Alpha adenovirus (AAV) | 7 mutations vs many | 70,002 / 12,581 | **0.137** | 0.281 | 0.355 | 0.404 | 0.297 | 0.421 |
| GB1[e] | Three vs many | 2,968 / 5,765 | **0.016** | 0.332 | 0.379 | 0.293 | 0.273 | 0.426 |
| GB1[e] | Two vs many | 427 / 8,306 | **0.030** | 0.442 | 0.427 | 0.107 | 0.124 | 0.461 |

Note that in all cases, to give the deep learning models the benefit of the doubt, we use the *best* performance for that uncertainty estimation technique, regardless of whether using one-hot or ESM embeddings as input. [a.] In many cases, there are multiple "splits" for the same dataset; for the AAV dataset, for example, the "designed vs mutant" split uses designed sequences as a training set and mutants as test, while the "mutant vs designed" split is the reverse.

**Table 4. Comparison with state of the art deep learning models for prediction of energetic and thermodynamic properties of small molecules, QM9 dataset**

| Model | U 0K MAE, kcal/mol | U 298K MAE, kcal/mol | H 298K MAE, kcal/mol | G 298K MAE, kcal/molf | Cv 298K, mol/cal * k | ZPVE, kcal/mol |
|---|---|---|---|---|---|---|
| MPNN[50] (deep learning) | 2.05 | 2.00 | 2.02 | 2.43 | 0.42 | |
| DTNN[50] (deep learning) | 2.43 | 2.43 | 2.43 | 2.02 | 0.27 | |
| Cormorant[51] (deep learning) | 0.484 | 0.507 | 0.484 | 0.461 | 0.026 | 0.047 |
| Enn-s2s[52] (deep learning) | 0.44 | 0.44 | 0.39 | 0.44 | 0.04 | 0.035 |
| SchNet[53] (deep learning) | 0.323 | 0.438 | 0.323 | 0.323 | 0.033 | 0.039 |
| NMP-EDGE[54] (deep learning) | 0.242 | 0.244 | 0.261 | 0.281 | 0.032 | 0.034 |
| PhysNet[55] (deep learning) | 0.188 | 0.189 | 0.194 | 0.217 | 0.028 | 0.032 |
| DIME[56] (deep learning) | 0.185 | 0.182 | 0.187 | 0.207 | 0.025 | 0.030 |
| Molformer[57] (deep learning) | 0.173 | 0.172 | 0.170 | 0.169 | 0.025 | 0.047 |
| HMGNN[58] (deep learning) | 0.137 | 0.158 | 0.14 | 0.175 | 0.023 | 0.027 |
| PAiNN[59] (deep learning) | 0.135 | 0.134 | 0.138 | 0.187 | 0.024 | 0.030 |
| | | | | | | |
| xGPR, GraphConv1d, 16,384 random features | 0.235 | 0.235 | 0.234 | 0.256 | 0.0260 | 0.0286 |
| xGPR, GraphConv1d, 32,768 random features | 0.183 | 0.183 | 0.182 | 0.204 | 0.0236 | 0.0269 |
| xGPR, GraphConv1d, 66,536 random features | 0.164 | 0.164 | 0.164 | 0.187 | 0.0223 | 0.0258 |

**Figure 1. Number of epochs (passes over the full dataset) required to fit**

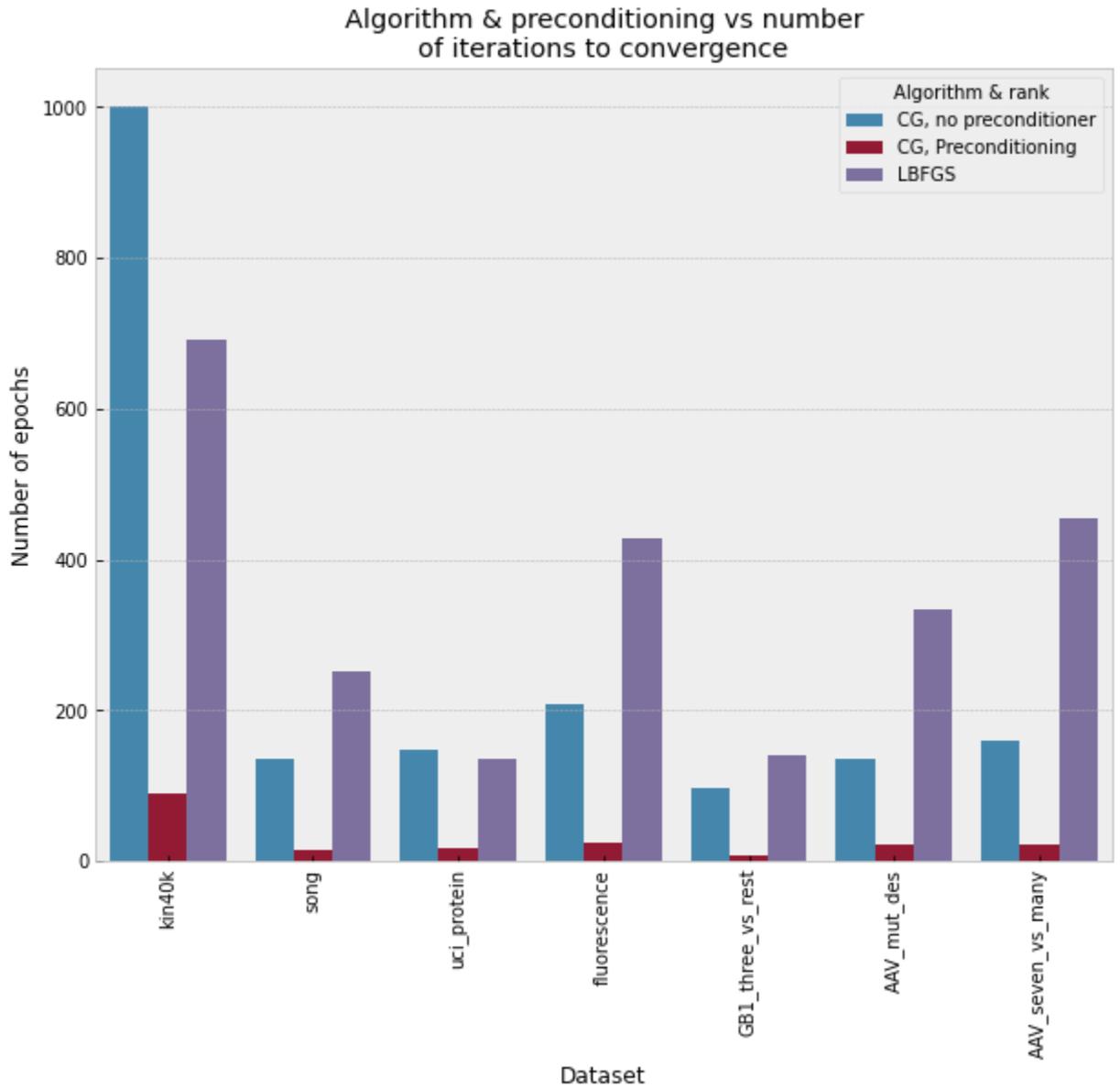

In all cases, a relative tolerance of 1e-6 is used as the threshold for convergence. For details on preconditioner rank, see Supporting Info S4.

**Figure 2. Relative error vs number of epochs: convergence rate of stochastic gradient descent and CG with and without preconditioning**

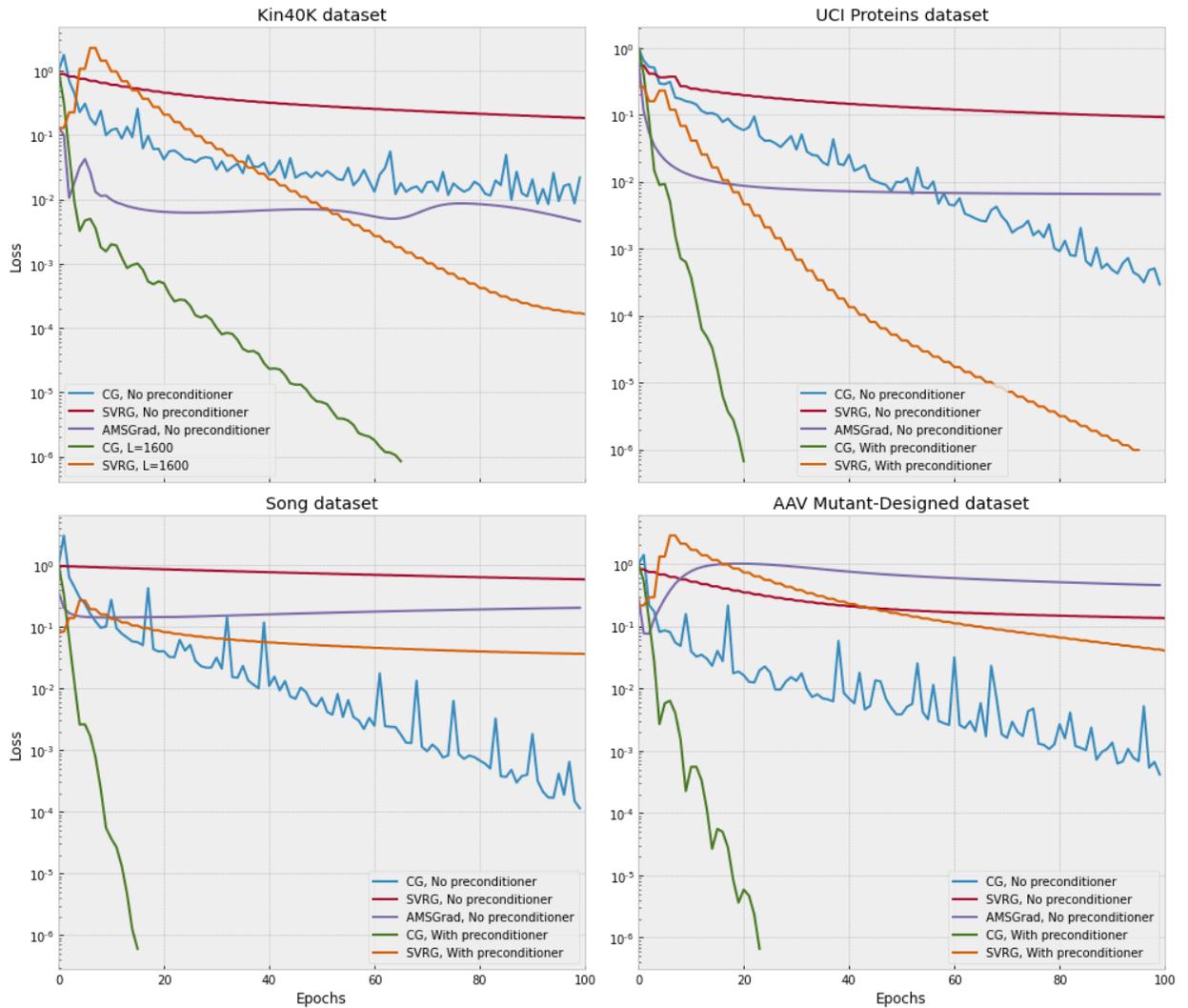

We use AMSGrad here since in initial experiments, the closely related Adam stochastic gradient descent method popular for fitting neural networks was even worse (failed to reduce error significantly in less than 100 epochs).

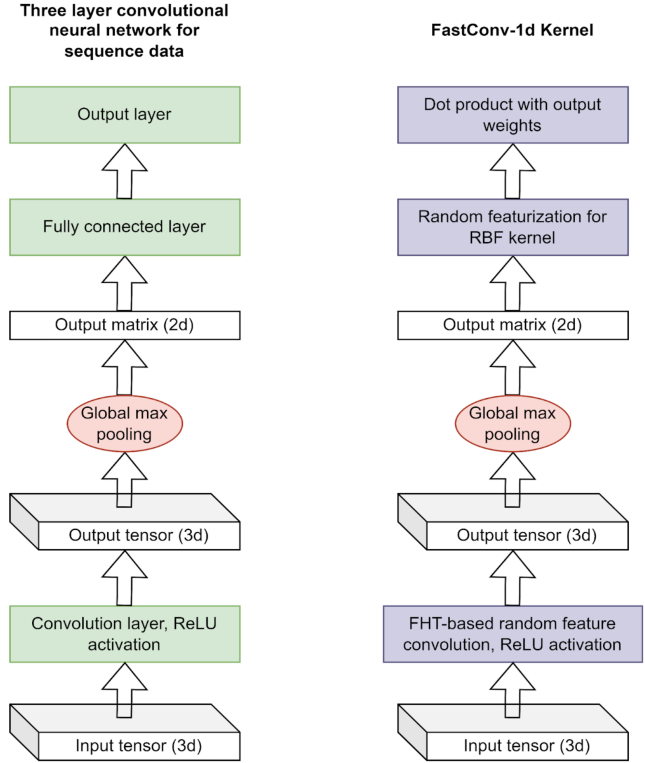

**Figure 3. Schematic of the Fast-Conv-1d kernel as compared with a standard 3-layer convolutional neural network.**

# SUPPORTING INFORMATION

## S1. Random features

We consider only positive definite kernel functions, and initially only stationary kernels, i.e. kernels for which $K(x_i, x_j) = f(x_i - x_j)$. We seek a transformation $\phi(x)$ such that the kernel $K(x_i, x_j) \approx \phi(x_i)\phi(x_j)$. From Bochner's theorem[1], any of the positive definite kernels we consider is the Fourier transform of a corresponding positive measure, i.e.:

$$K(x_i, x_j) = \int p(\omega) e^{i\omega^T(x_i - x_j)} d\omega$$

We can approximate this expectation using Monte Carlo sampling, i.e.:

$$K(x_i, x_j) \approx \frac{2}{M}\sum_k^M \cos(\omega_k^T x_i)\cos(\omega_k^T x_j) + \sin(\omega_k^T x_i)\sin(\omega_k^T x_j)$$

Since Monte Carlo approximation is an unbiased estimator of the mean, our estimate converges on the true value as we increase the number of random features. The variance of this estimator is proportional to $(1 / M)^2$.

Although this procedure was originally derived in the context of stationary kernels, we can apply a similar Monte Carlo sampling approach to certain nonstationary kernels, most importantly a class of kernels dubbed "neural network kernels" and also called "pointwise nonlinear Gaussian kernels" in more recent literature[3,4]. Consider a neural network with one hidden layer and one output layer:

$$f(x) = \sum_i^{N_H} v_i h(Wx + b)_i$$

where x is an input vector, W is a learned weight matrix of dimensions D x $N_H$ where D is the dimensionality of the input, $N_H$ is the number of units in the hidden layer, b is an offset or bias vector of length $N_H$, h is some activation function and $v_i$ is a learned weight. Assume that all weights in W are iid drawn from $N(0, \sigma_w^2)$, the weights $v_i$ are iid drawn from $N(0, \sigma_v^2)$ and that the bias terms are iid drawn from $N(0, \sigma_b^2)$. If we augment each x with an additional element that is 1 for all data points, we can rewrite this more simply as:

$$f(x) = \sum_i^{N_H} v_i h(Wx)_i$$

Where each row of W is drawn iid from $N(0, \Sigma_W)$, when $\Sigma_W$ is a diagonal covariance matrix whose first diagonal element is $\sigma_b^2$ and all remaining diagonal elements are $\sigma_w^2$.

Let S denote the collection of all weights; it follows that

$$\Sigma_S[f(x_1)f(x_2)] = N_H \sigma_v^2 \Sigma_W[h(Wx_1)_i h(Wx_2)_i]$$

We see that we can once again approximate the expectation term on the right hand side using Monte Carlo sampling and thereby obtain a formulation very similar to the one for stationary kernels. This kernel is only positive definite for certain choices of activation function $h$[3]. One such choice is the ReLU activation function which yields the order 1 arccosine kernel[5], although the error function has also been suggested as a useful activation function[3]. Interestingly, the order 1 arccosine kernel as formulated in the literature omits the bias terms. During the implementation of our library, we found that the arccosine kernel performs very poorly without the bias term and requires the inclusion of the bias term to achieve acceptable performance.

For both stationary and neural network kernels, then, we can implement the random Fourier features approach as follows. If X is an N x D matrix for N data points and D input features, we populate a D x M matrix with M samples from $p(\omega)$, where $p(\omega)$ is the Fourier transform of the kernel for stationary kernels and the normal distribution for neural network kernels. We multiply this sample matrix by X, and for neural network kernels add an additional bias vector to the result. We then apply the appropriate activation function (e.g. sine and cosine for stationary kernels) and multiply by a constant to yield an N x M matrix Z (or, for stationary kernels, N x 2M). The approximate kernel matrix is then given by $ZZ^T$.

The distribution across predicted values $f_*$ at new datapoint $x_*$ for a mean-zero Gaussian process given training dataset X with corresponding ground truth values *y* is:

$$p(f_* | X, y) \sim N(\mu_*, \Sigma_*)$$
$$\mu_* = K(x_*, X)[K(X, X) + \lambda^2 I]^{-1} y, \qquad \sigma_* = K(x_*, x_*) - K(x_*, X)[K(X, X) + \lambda^2 I]^{-1} K(X, x_*)$$

Since we approximate the kernel as $ZZ^T$, we can use the matrix inversion lemma and the Woodbury matrix identity to rewrite these. The mean for example becomes:

$$\mu_* = z_*^T [Z^T Z + \lambda^2 I]^{-1} Z^T y$$

where Z is the training set transformed as described above, I is the identity matrix, $\lambda$ is a hyperparameter which ensures the kernel matrix is positive definite and z is the new datapoint transformed as described above. If we consider

$$[Z^T Z + \lambda^2 I]^{-1} Z^T y$$

to be the weights, the predictive mean corresponds to linear regression in the feature space corresponding to the random feature map. It is easy to see the correspondence between a neural network with a single hidden layer plus an output layer and the above kernels approximated using this formulation, although there are several key differences. The approximate GP "learns" by changing the distribution of the hidden layer weights, not individual values, and can be fitted by maximizing the marginal log-likelihood rather than the likelihood; also, the marginal likelihood and the posterior predictive distribution are available in closed form. It is also possible to construct kernels which are analogous to neural networks with several hidden layers. In the main text, we introduce the Fast-Conv-1d kernel, which corresponds to a three-layer neural network with a convolution layer, a fully-connected layer and an output layer, all populated with random weights; this concept is also illustrated in Figure 3 of the main text.

Note that since the error in the approximation decays exponentially with an increasing number of random features, we do not generally need a large number to obtain a good approximation; indeed, increasing random features past 16,000 or so yields sharply diminishing returns, and 8,192 is sufficient for many tasks. For predicting the mean $\mu_*$, since this is the most important quantity, practitioners may want to use a large number of random features to "buy" small improvements in accuracy, but for quantifying variance 512 - 1,024 is generally sufficient in our experience. Consequently, xGPR allows users to use a smaller number of random features to quantify variance than for predicting the mean (to reduce the computational expense of predictions for variance).

## S2. Orthogonal random features

Consider the case where the number of random features *D* is the same as the dimensionality of the input, *d*, and where we multiply the input vector *x* by a matrix M of random features drawn iid from a normal distribution.

If we would like the rows of M to be orthogonal, we can first take the QR decomposition of M, then multiply the resulting Q by a diagonal matrix S whose diagonal elements are drawn from a $\chi$-distribution with *d* degrees of freedom. The input vector *x* is now transformed by computing the product $kSQx$. If the number of random features is greater than *d*, we can repeat this procedure *m* times where *m* is the result of ceiling division of D by *d*, and then discard any features in excess of D. If D is less than *d*, we can generate *d* random features and discard any in excess.

For certain kernels, in particular the squared exponential kernel, it has been shown that fewer orthogonal features are required to accurately approximate the kernel[4], so we achieve a more accurate approximation for the same number of random features by using orthogonal random features. In addition to this benefit, the orthogonal feature approach can be further modified to realize a large reduction in computational cost. Yu et al.[6] show that we can achieve nearly equivalent results by replacing the matrix Q with the following product:

$$HD_1 HD_2 HD_3$$

where H is the normalized Hadamard matrix and each D is a separate diagonal matrix whose entries are drawn from the Rademacher distribution. The memory footprint of a fitted model is greatly reduced, because we need to only store the diagonal elements of the D-matrices. More importantly, similar to the substitution of the fast Fourier transform for multiplication with a matrix in the discrete Fourier transform, we substitute the fast Hadamard transform for the Hadamard matrix multiplication, thereby reducing the cost of this procedure to O(D log D) – a dramatic benefit for large M. The general procedure for transforming the input is shown for the RBF kernel in Algorithm S1 (other kernels similar).

Note that Yu et al.[6] suggest replacing the S-matrix used here with multiplication by a constant, $\sqrt{d}$; we have found this to fail on 1d data, where the resulting kernel is unable to provide an acceptable fit for simple toy problems, and it provides no benefit on higher dimensional data. Our version from above does not suffer from this limitation.

---

**Algorithm S1** Orthogonal random features transform for RBF kernel

---

**Kernel object initialization**
**Input:** Expected dimensionality *d* of input; expected number of sampled frequencies *M*
1. $D = 2^{ceiling(log_2(max(d, 2)))}$
2. If D < M:
    a. $P = ceiling(M / D)$
3. Else:
    a. $P = 1$

---

4. Create, store diagonal matrix S with size *M x M* with elements drawn from the χ-distribution with degrees of freedom *D*
5. For i in 1…P:
   a. Create, store diagonal matrices $D_{ai}$, $D_{bi}$, $D_{ci}$ with size *D* with elements drawn from the Rademacher distribution

**Kernel applied to transform input data**
**Input:** *N x d* matrix $x_i$, which is a chunk of the input data; kernel hyperparameters β, σ
**Output:** kernel-transformed matrix $Z_i$

1. For i in 1…P:
   a. Create empty *N x D* array $Y_i$
   b. Copy $x_i$ into the first *d* columns of $Y_i$; set remaining elements to 0
   c. For k in [a, b, c]:
      i. $Y_i = H(Y_i D_{ki})$, where H is a normalized in-place fast Hadamard transform
2. Concatenate $Y_1 ... Y_P$ to form *N x D\*P* array Y
3. If *D\*P > M*:
   a. Discard columns *M…D\*P* of Y
4. $Y = \sigma Y S$
5. $Y_1 = cos(Y)$, $Y_2 = sin(Y)$   where sin and cos are applied elementwise
6. Concatenate $Y_1$ and $Y_2$ to form *N x 2M* matrix Z
7. $Z = \beta \sqrt{\frac{2}{M}} Z$
8. Return Z

**S3. Performance of the fast Hadamard transform**
We implement a Python-wrapped C / Cuda implementation of the fast Hadamard transform for CPU and GPU. We compare the speed of our implementation with matrix multiplication against a matrix of random features as implemented in the Numpy library (v 1.22.3) and Cupy library for GPU (v 10.4.0) – see Table S2 and S3 for the results. Briefly, our implementation is faster than matrix multiplication for modest numbers of random features, and the performance advantage of our implementation increases with larger numbers of random features as expected, enabling us to quickly scale to large numbers of random features. To compare with other fast transforms, we also compare with the discrete cosine transform (DCT) on CPU as implemented in the Scipy library (v 1.8.0) (Table 1 in the main text).

The results in milliseconds appear in Table S1 and Table S2. In Table S1, we evaluate the time needed to generate random features if the input matrix has as many columns as there are features – this is a somewhat unusual situation but provides a dramatic demonstration of the speed increase for the fast Hadamard transform-based procedure. In Table S2, we consider the more realistic situation where there are a fixed number of 1024 input features and we must generate an increasing number of random features.

We use an input matrix with 2000 rows since this chunk size is common when we work on real datasets. For the GPU implementation, we use 32-bit (single precision) floats due to the substantially greater performance for single precision matrix multiplication on most GPUs, while for the CPU tests we use 64-bit (double precision). In general, however, unless the user selects otherwise, xGPR performs random feature generation using single precision for both CPU and GPU.

**Table S1: Speed of the SORF operation implemented in this study (highlighted in the table) compared with vanilla random features and with a generic Scipy fast transform**

| Algorithm | Average time (ms), 2000 x 2048 input matrix | Average time (ms), 2000 x 4096 input matrix | Average time (ms), 2000 x 8192 input matrix | Average time (ms), 2000 x 16384 input matrix |
|---|---|---|---|---|
| Matrix multiplication, CPU, Numpy, float64 | 110 | 450 | 1700 | Not tested (cost prohibitive) |
| **SORF transform, 2 threads, CPU, float64** | 30 | 70 | 160 | 350 |
| Scipy DCT (3x), float64 | 80 | 190 | 390 | 880 |
| **SORF transform, GPU, float32** | 6 | 13 | 30 | 60 |
| Matrix multiplication, GPU, Cupy, float32 | 3.5 | 13 | 50 | 210 |

*The results shown here are rounded to reflect precision, include both CPU time and GPU time (where applicable) and are calculated as the average time per operation for 100 repeats of the specified operation. All experiments were performed on an Intel i5-7500 CPU and a GeForce GTX1070 GPU. The methods used in this study are in bold.*

**Table S2: Speed of the SORF operation implemented in this study (highlighted in the table) compared with vanilla random features and with a generic Scipy fast transform**

| Algorithm | Average time for 2048 features (ms), 2000 x 1024 input matrix | Average time for 4096 features (ms), 2000 x 1024 input matrix | Average time for 8192 features (ms), 2000 x 1024 input matrix |
|---|---|---|---|
| Matrix multiplication, CPU, Numpy, float64 | 56 | 121.6 | 236.6 |
| **SORF transform, 2 threads, CPU, float64** | 29.2 | 58.5 | 127.6 |
| Scipy DCT (3x), float64 | 65.5 | 145.6 | 286.7 |
| **SORF transform, GPU, float32** | 1.344 | 2.75 | 5.1 |
| Matrix multiplication, GPU, Cupy, float32 | 1.77 | 3.533 | 6.56 |

*The results shown here are rounded to reflect precision, include both CPU time and GPU time (where applicable) and are calculated as the average time per operation for 100 repeats of the specified operation. All experiments were performed on an Intel i5-7500 CPU and a GeForce GTX1070 GPU. The methods used in this study are in bold.*

**S4. Preconditioning for conjugate gradients and stochastic gradient descent**

Consider the well-known conjugate gradients algorithm, which solves the system $Aw = b$ for $w$. It can be shown[7] that the residual error on iteration $n$ has as its upper bound:

$$2 \left(\frac{\sqrt{\kappa}-1}{\sqrt{\kappa}+1}\right)^n \|e_0\|$$

where $\kappa$ is the condition number of A, and $\|e_0\|$ is the error of the initial guess. If $\kappa$ is large, the algorithm will therefore converge very slowly. A preconditioner whose inverse $M^{-1}$ is a close approximation to the inverse of A will reduce the condition number of A and accelerate convergence[8]. It is critical for our purposes to have a preconditioner that can be constructed without explicitly forming the matrix $Z^T Z$, which will require expensive matrix multiplications and for large numbers of random features will be too large to store in memory.

Kernel matrices for Gaussian processes have frequently been approximated using the Nystrom method, which provides the following low-rank approximation:

$$A \approx A_{m,q} A_q^{-1} A_{m,q}^T$$

where m is the number of rows in A, and q is a randomly selected subset of the columns of A. The Nystrom approximation is cheap to construct and has worked well for many problems in practice[9]. Although sampling from a uniform distribution has traditionally been the most popular approach, the performance of this method is sensitive to the sampling method used to select columns and rows in A[10].

We found the randomized Nystrom approximation described by Alaoui et al.[11] to provide better performance. In this scheme, for the system:

$$(Z^T Z + \lambda^2 I)^{-1} Z^T y = w$$

A low-rank approximation of $Z^T Z = A$ is provided by:

$$A_{nystrom} \approx (A\Omega)(\Omega^T A\Omega)^\dagger (A\Omega)^T$$

where for $\Omega$ we use an iid matrix drawn from a zero mean, unit-variance normal distribution; $\Omega$ is of size M x L, where M is the number of random features. This approach provides a rank-L approximation to A, and a numerically stable approach for generating it (which we implement for our library) is described in Martinsson et al.[12] and appears in Algorithm S2. Importantly for our purposes, we can construct this as shown *without* ever explicitly forming the matrix $Z^T Z$ and using a single loop over the dataset, loading one chunk of the data into memory at a time.

---

**Algorithm S2** Constructing a randomized Nystrom preconditioner

---

**Input:** Dataset X stored in chunks $x_1 ... x_n$ on disk, desired preconditioner size L, GP kernel, number of random features *M*

**Output:** Factored preconditioner in the form $U\Lambda U^T$, where U is orthonormal and $\Lambda$ is diagonal

6. Generate *M x L* matrix $\Omega$ by sampling from $N(0, 1)$
7. $\Omega = qr(\Omega)$                          QR decomposition
8. Generate *M x L* matrix *Q* and initialize to all zeros
9. For i in 1…n, load $x_i$
    a. Use Algorithm S1 to generate $z_i$
    b. $Q \mathrel{+}= z_i^T(z_i\Omega)$
10. $\nu = \sqrt{M}eps(norm(Q))$
11. $Q_\nu = Q + \nu\Omega$                         For numerical stability
12. $C = cholesky(\Omega^T Q_\nu)$
13. $CB = Q_\nu$                               Triangular solve for B
14. $U, \Sigma, \sim = svd(B)$                       SVD of B
15. $\Lambda = max(0, \Sigma^2 - \nu I)$
16. Return $U, \Lambda$

We here introduce a novel modification. We replace the M x L matrix $\Omega$ with the subsampled randomized Hadamard transform, which is defined as follows:

$$SRHT = \sqrt{\frac{M}{L}}SHD$$

Where *D* is a diagonal matrix with entries drawn from the Rademacher distribution, H is the normalized Hadamard matrix, and S is a subset of L rows drawn with equal probability from the identity matrix. In practice, of course, we use the fast Hadamard transform rather than the Hadamard matrix. For step 9b, we then substitute:

$$Q \mathrel{+}= SRHT(Z^T)Z$$

In step 12, for the product $\Omega^T Q_\nu$, we substitute $SRHT(Q_\nu)$. Finally, to ensure numerical stability, we replace the shift-then-subtract procedure above (steps 11 and 15) with a procedure outlined by Li et al.[13], wherein rather than taking the Cholesky decomposition of $SRHT(Q_\nu)$ in step 12, we find the self-adjoint square root of this matrix using SVD, then use this to solve for B in the following step.

Using the SRHT to construct low-rank matrix approximations has been suggested before (see Halko et al.[14] and Boutsidis et al.[15] for example), but not to construct a preconditioner for CG / stochastic gradient descent as we suggest here (at least not to our knowledge). It significantly reduces the cost of preconditioner construction, since we replace a matrix multiplication with $O(NML)$ scaling with a fast transform operation; this is especially advantageous if working on CPU. Empirically, we have found the SRHT-based preconditioner construction routine to be substantially faster even for datasets of only 40,000 datapoints, and the advantage increases with increasing dataset size and number of random features.

In every case we encountered, we found empirically that a preconditioner constructed using SRHT as we suggest here performed either as well as or *better than* a preconditioner constructed using the unmodified algorithm S2. In Figure S1 below, we compare the number of iterations required for conjugate gradients to converge to a pre-specified threshold using a preconditioner constructed either with SRHT or with the unmodified algorithm S2, using a variety of different values for *L*. We see that in every case, the SRHT-built preconditioner achieves the same or greater acceleration, and indeed this has been consistently true throughout our experiments. For the time being, we provide both the modified and unmodified preconditioner construction algorithms as options the user can select in xGPR and default to the SRHT-based algorithm in light of its superior speed and empirical performance. A theoretical analysis of the merits of each approach is beyond the scope of this paper, given the many other problems we are already tackling; we defer such an analysis to future work.

**Figure S1. A comparison of the acceleration achieved for conjugate gradients using a preconditioner constructed using SRHT or using the unmodified Algorithm S2.**

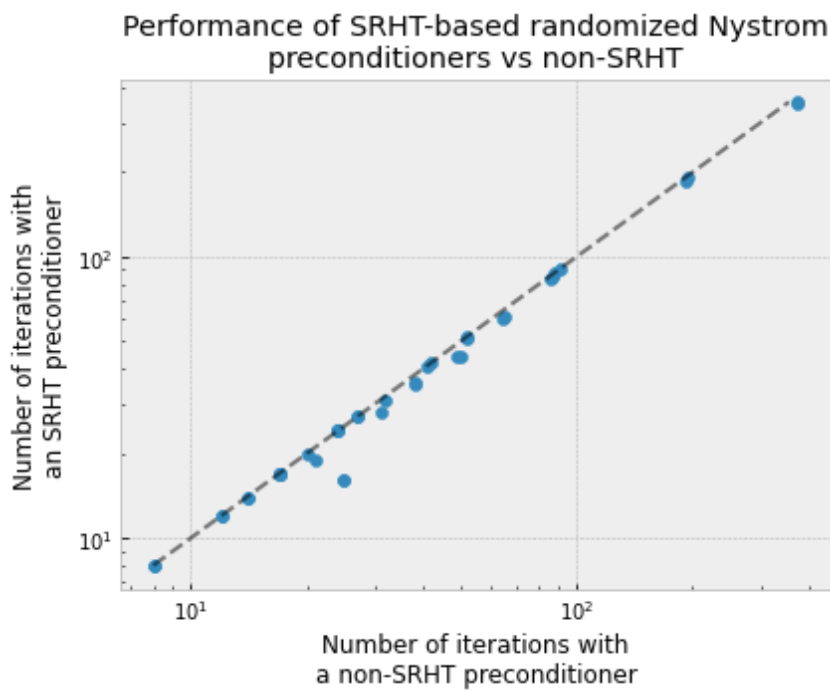

Additionally, we can further modify the preconditioner to improve its performance. In step 9 of Algorithm S2, we form the product $Z^T(Z\Omega)$, where Z is the matrix of random features for the training data of shape N x M for M random features and N datapoints. We can retrieve the matrix Q from the QR decomposition of this product and use this to form a second product $Z^T(ZQ)$ using a second pass over the dataset. We then use this product in Algorithm S2 in place of the matrix Q from step 9. (We can even repeat this process a third time, although this offers a negligible improvement in preconditioner performance.) As expected, the resulting preconditioner reduces the number of iterations required for conjugate gradients to converge in all of our experiments, in most cases by 20-25%, as illustrated in Figure S2 below. It is however more expensive to construct since it requires two passes over the dataset. Whether this is beneficial will depend on the size of the dataset and on available hardware. We therefore allow the user to select "srht_2" as an option for preconditioner construction in xGPR and provide guidance in the documentation on when to choose this option.

**Figure S2.** The number of iterations to convergence across various benchmark datasets using various settings for preconditioner rank and three different variants on the preconditioner construction algorithm.

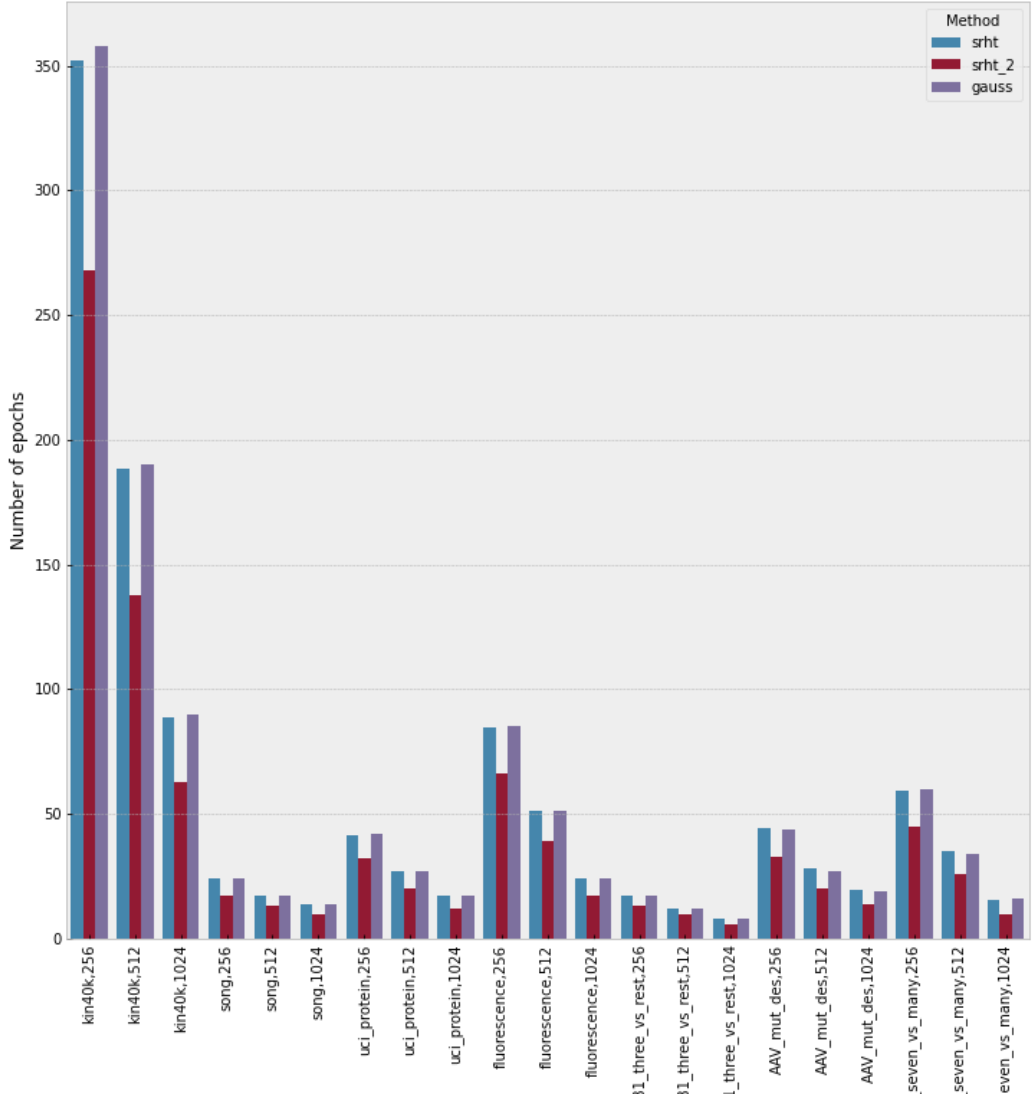

Irrespective of which option is used to construct it, the inverse of the preconditioning matrix takes the form:

$$M^{-1} = (\beta_L + \lambda^2)U(\Lambda + \lambda^2 I)^{-1}U^T + (I - UU^T)$$

Where $\beta_L$ is the smallest eigenvalue of the preconditioner and $\lambda$ is a hyperparameter of the GP kernel. We do not of course actually form the matrix $M^{-1}$ explicitly, but since we have generated all the needed values above, we can supply the matrix-vector product $M^{-1}v$ on demand, which is all that is required for preconditioned conjugate gradients or preconditioned stochastic gradient descent.

The larger L, the better the approximation; if $\beta_1 \geq \beta_2 \geq \beta_3...$ are the eigenvalues of A, the approximation is also better the more rapidly the eigenvalues decay, and the smaller the eigenvalues for $\beta_{j \geq L}$. Let $U\Lambda U^T$ be the eigendecomposition of $A_{nystrom}$ of rank L, let the preconditioner P be $P = \frac{1}{\beta_L + \lambda^2} U(\Lambda + \lambda^2 I) U^T + (I - UU^T)$, and let $\beta_L$ be the smallest eigenvalue of $A_{nystrom}$. It can be shown[16] that if $\kappa_2$ is the condition number of the preconditioned system, the probability that:

$$max(0, \kappa_2(P^{1/2}(\lambda^2 I + Z^T Z) P^{1/2}) - \frac{\beta_L + \lambda^2}{\lambda^2}) \leq \frac{\beta}{\delta}$$

will be $\geq 1 - \delta$, where $\delta$ is some failure tolerance.

How to choose L? The value of $\frac{\beta_L}{\lambda^2}$ for a given L can be estimated by constructing a preconditioner, and the value of L can then be doubled until $\frac{\beta_L}{\lambda^2}$ suggests the number of iterations required to fit the model is likely to be acceptable. As illustrated in Figure 1a from the main text, the log of the number of iterations often exhibits a roughly linear relationship with $\frac{\beta_L}{\lambda^2}$ for a given desired tolerance. Consequently, we can estimate the number of iterations required to fit for a given $\frac{\beta_L}{\lambda^2}$ and if we consider the resulting number of iterations to be too high, we can adjust L and reconstruct the preconditioner. For more explicit guidance on how to choose a preconditioner, refer to the xGPR documentation at https://xgpr.readthedocs.io/en/latest/ .

The relationship between the value of L and the number of iterations required to fit using CG is illustrated in Figure S below.

**Figure S3.** Number of epochs / iterations required for convergence with CG on different datasets as a function of beta / lambda^2.

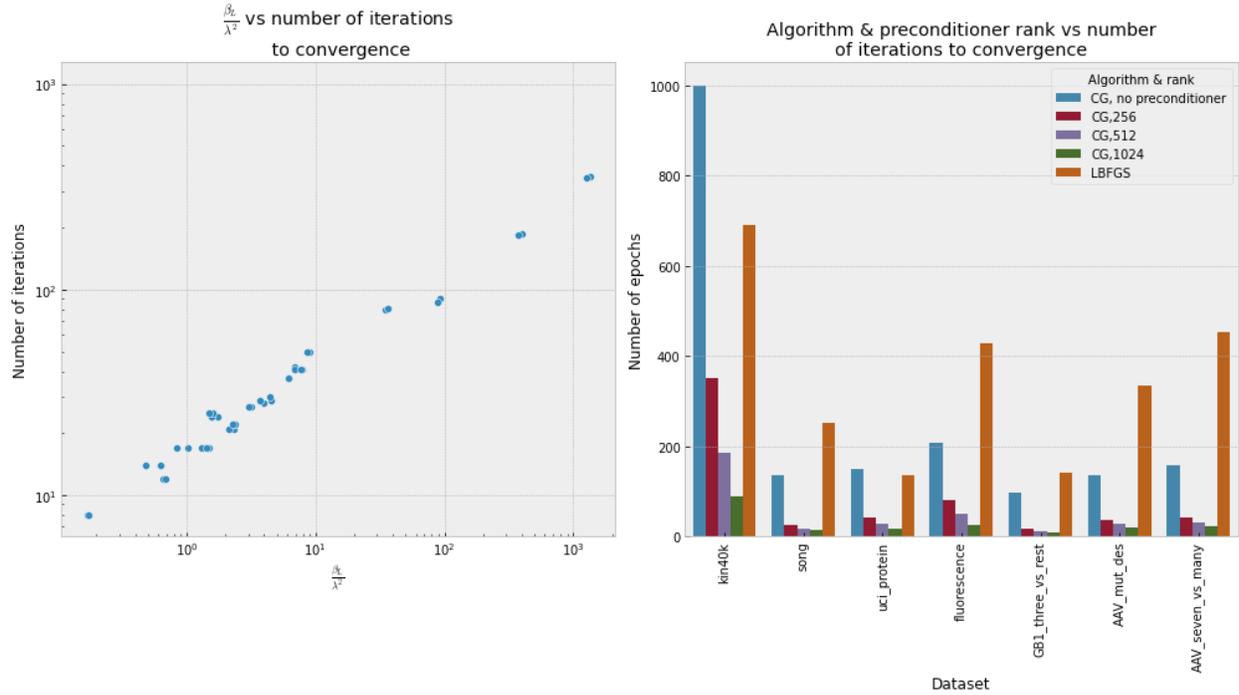

The preconditioner enables us to quickly fit the model but does not address the issue of hyperparameter tuning. Calculating the marginal likelihood of a Gaussian process requires calculating the log determinant of an M x M matrix for M random features, which will be very expensive if M is large. We therefore develop an approach for estimating marginal likelihood for large numbers of random features.

## S5. Log determinant approximation and conjugate gradients

Hyperparameters can be tuned for a GP using performance on a validation set (as for any other model), and this approach is implemented as an option in xGPR, but it is also possible to tune hyperparameters by maximizing the marginal likelihood. This procedure is more robust to overfitting and also means that we do not need a validation set, so we prefer to use this approach where possible. Nonetheless, tuning using marginal likelihood does present one particular challenge. Recall that for a Gaussian process we find the following for the negative marginal log likelihood:

$$-\log p(y | X) = -\log \int p(y|f, X) p(f | X) df = \frac{1}{2} y^T K_y^{-1} y + \frac{1}{2} \log |K_y| + \frac{N}{2} \log(2\pi)$$

Where $K_y = K(X, X) + \lambda^2 I$. In the random features scheme, we substitute our approximation $Z^T Z$ for $K(X, X)$ and rearrange to yield:

$$-\log p(y | X) = \frac{-1}{2\lambda^2} y^T Z (Z^T Z + \lambda^2 I)^{-1} Z^T y + \frac{1}{2\lambda^2} y^T y + (n - m)\log(\lambda) + \frac{1}{2} \log |Z^T Z + \lambda^2 I| + \frac{n}{2} \log(2\pi)$$

As we have shown above under S4, we can efficiently compute $(Z^T Z + \lambda^2 I)^{-1} Z^T y$ without ever forming the matrix $(Z^T Z + \lambda^2 I)$ explicitly using preconditioned conjugate gradients. The $log|Z^T Z + \lambda^2 I|$ however is more challenging to compute if the number of random features is large. Under section S6 we will develop some highly efficient ways to tune hyperparameters with a small number of random features, e.g. < 4000, which we frequently do in practice. First however we will introduce a method to use preconditioned conjugate gradients to approximate $log|Z^T Z + \lambda^2 I|$ without ever forming the $Z^T Z + \lambda^2 I$ matrix explicitly. This will enable us to use the same tool we use for fitting to tune hyperparameters in an efficient way even if the number of random features is large.

A variety of approaches have been outlined in the literature for approximating the log determinant of a large matrix $A$, including stochastic Lanczos quadrature[17], stochastic trace estimation combined with Chebyshev polynomials[18], and subspace iteration[19]. The subspace iteration method of Saibaba et al. is only appropriate if a small subset of the eigenvalues of $A$ are dominant, i.e. there is a large eigengap with eigenvalue $\lambda_i >> \lambda_{i+1}, \lambda_{i+2}...\lambda_N$, which is not always true during tuning.

Stochastic Lanczos quadrature (SLQ) is especially attractive in this context, since for large numbers of random features we use preconditioned conjugate gradients to fit the model, and we can automatically generate the tridiagonal matrices required for stochastic Lanczos quadrature during the course of the conjugate gradients optimization at negligible additional cost (see Saad 2003[7]). Gardner et al.[20] have already demonstrated the use of this approach to accurately estimate log determinants for exact (non-approximated) Gaussian processes using an incomplete Cholesky decomposition to build the preconditioner.

In Algorithm S5 below, we show how to use this approach in combination with the random features approximation and the randomized Nystrom approximation to accurately estimate the log determinant term, $\frac{1}{2}log|Z^T Z + \lambda^2 I|$, while simultaneously calculating the performance penalty $\frac{-1}{2\lambda^2} y^T Z (Z^T Z + \lambda^2 I)^{-1} Z^T y$ (the remaining terms in the marginal likelihood are either constants or do not require us to retrieve the data and hence are easily calculated). The resulting algorithm exhibits linear scaling with dataset size, subquadratic scaling with number of random features, and never requires us to load more than one chunk of data into memory at a time.

As a preliminary, we first rewrite the log determinant and make use of Hutchison's trace estimator[21]:

$$log|Z^T Z + \lambda^2 I| = trace(log(Z^T Z + \lambda^2 I)) \approx \frac{1}{n_v} \sum_{i=1}^{n_v} p_i^T log(Z^T Z + \lambda^2 I) p_i$$

Where $p_i$ are probe vectors drawn(typically) from either a Rademacher or a Gaussian distribution.

Next, we note that when we use preconditioned conjugate gradients with preconditioner $P$, we are in fact solving the preconditioned system $log|P^{-1/2} Z^T Z + \lambda^2 I\, P^{-1/2}|$. We can adopt the same approach that Gardner et al.[20] adopt for exact (un-approximated) GPs and rewrite this as:

$$\log|Z^T Z + \lambda^2 I| = \log| P^{-1/2} (Z^T Z + \lambda^2 I) P^{-1/2} | + \log|P|$$

$$\log| P^{-1/2} (Z^T Z + \lambda^2 I) P^{-1/2} | = trace(P^{-1/2} \log(Z^T Z + \lambda^2 I) P^{-1/2}) \approx$$

$$\frac{1}{n_v} \sum_{i=1}^{n_v} p_i^T P^{-1/2} \log(Z^T Z + \lambda^2 I) P^{-1/2} p_i$$

We therefore need to apply the preconditioner to the probe vectors. This is easily done if the probe vectors are drawn from a Gaussian; we can merely draw them from a normal distribution whose covariance matrix is our preconditioner. We now refer to these preconditioned probe vectors as $c_i$.

We now need to evaluate the term

$$\frac{1}{n_v} \sum_{i=1}^{n_v} c_i^T P^{-1/2} \log(Z^T Z + \lambda^2 I) P^{-1/2} c_i$$

We do so using the quadrature-based procedure of Ubaru et al.[17], but using the preconditioned conjugate gradients algorithm; by storing the alpha and beta coefficients produced on each iteration of conjugate gradients, we generate the same tridiagonal matrix that Ubaru et al. construct via the Lanczos algorithm. We merely need in this case to run conjugate gradients with multiple target vectors – both the $Z^T y$ vector so that we can calculate the performance penalty $(Z^T Z + \lambda^2 I)^{-1} Z^T y$, and the preconditioned probe vectors so that we can generate $n_v$ tridiagonal matrices $T_1, T_2 ... T_{n_v}$. A simplified version of the full procedure is outlined in Algorithm S5 below.

**Algorithm S5** Preconditioned conjugate gradients for simultaneous estimation of $(Z^T Z + \lambda^2 I)^{-1} Z^T y$ and $\log|Z^T Z + \lambda^2 I|$

**Input:** Dataset X with N datapoints stored in chunks $x_1 \ldots x_n$ and $y_1 \ldots y_n$ on disk, initialized kernel with fixed hyperparameters that generates length-$M$ random feature representation $k(x_i)$ for input datapoint $x_i$, preconditioner $P$ constructed using Algorithm S2 or appropriate variant, number of probe vectors $n_v$, convergence threshold *tol*

1. Create empty array $Z^T y$ of shape $M$
2. For i in 1,2…n:
    a. $Z^T y \mathrel{+}= k(x_i)^T y_i$
3. For i in 1,2…$n_v$:
    a. Draw probe vector $c_i$ from $N(0, P)$
4. Create empty array $r_j$ of shape $(M, n_v+1)$
5. Populate the first $n_v$ columns of $r_j$ with $c_1, c_2, c_3 \ldots c_{n_v}$. Copy $Z^T y$ into the remaining column.
6. Let $L_1 = [||r_{j1}||, ||r_{j2}||, ||r_{j3}|| \ldots ||r_{jn_v+1}||]$ where $||r_{jk}||$ is the Euclidean norm of column $k$ in $r_j$, such that $L_1$ is a row vector of length $n_v + 1$
7. Create initial weight matrix $w_j$ of shape $(M, n_v+1)$; set all entries to zero.
8. Create empty array $B$ of shape $(M, n_v+1)$. Let $1_M$ be a length-M vector of ones. Let $err$ be a length $n_v+1$ vector of ones.
9. $p_j = z_j = P^{-1} r_j;\quad j = 1$
10. While $max(err) >$ tol and $j <$ maxiter:
    a. Set all entries of B to zero
    b. For i in range 1,2…n:
        i. $B \mathrel{+}= k(x_i)\, k(x_i)^T p_k$
    c. $B \mathrel{+}= \lambda^2 p_j$
    d. $\alpha_j = \dfrac{1_M^T (r_j \circ z_j)}{1_M^T (p_j \circ B)}$ with elementwise division, such that $\alpha_j$ is a row vector of length $n_v+1$
    e. $w_j \mathrel{+}= (1_M \alpha_j) \circ p_j$

f. $r_{j+1} = r_j - (1_M \alpha_j) * B$

g. Let $err = [||r_{j1}||, ||r_{j2}||, ||r_{j3}||.... ||r_{jn_v+1}||]$ where $||r_{jk}||$ is the Euclidean norm of column $k$ in $r_j$, such that $err$ is a row vector of length $n_v + 1$ where $||r_{ji}||$ is the norm of column $i$ in $r_j$

h. $err = err / L_1$    with elementwise division

i. $z_{j+1} = P^{-1} r_j$

j. $\beta_j = \frac{1_M^T (r_{j+1} \circ z_{j+1})}{1_M^T (r_j \circ z_j)}$    with elementwise division, such that $\beta_j$ is a row vector of length $n_v + 1$

k. $p_{j+1} = z_{j+1} + (1_M \beta_j) \circ p_j$

l. Store $\beta_j, \alpha_j$

m. $j \mathrel{+}= 1$

11. Let $\alpha_{ik}$ be the element of $\alpha_k$ corresponding to iteration $k$ and probe vector $i$ (corresponding to column $i + 1$ of $p_j$). Let $\beta_{ik}$ be the element of $\beta_k$ corresponding to iteration $k$ and probe vector $i$ (corresponding to column $i + 1$ of $p_j$).\

12. $\Gamma = 0$

13. For $i$ in $1,2...n_v$:

   a. Create empty matrix $T$ of shape $(j, j)$. Set all entries to zero.
   
   b. For $k$ in $1,2...j$:
   
   i. $T_{kk} = \frac{1}{\alpha_{ik}}$
   
   c. For $k$ in $1,2...j-1$:
   
   i. $T_{k+1,k+1} \mathrel{+}= \frac{\beta_{ik}}{\alpha_{ik}}$
   
   ii. $T_{k,k+1} = T_{k+1,k} = \frac{\sqrt{\beta_{ik}}}{\alpha_{ik}}$
   
   d. $[Y, \Theta] = Eig(T); \quad \tau_m = e_1^T y_m \quad for\ m\ in\ 1, 2... j$
   
   e. $\Gamma \mathrel{+}= \sum_m^j \tau_m log(\theta_m)$

14. $\Gamma = \Gamma * \frac{M}{n_v}$

15. Return estimated log determinant $\Gamma$, vector $Z^T y$ and column 1 of $w_j$; this last corresponds to $(Z^T Z + \lambda^2 I)^{-1} Z^T y$ and can be used to calculate the performance term of the negative marginal log likelihood

To evaluate the accuracy of this procedure, we randomly select 8 hyperparameter combinations each for 7 different datasets ranging in size from a few hundred datapoints to half a million. For each, we calculate the negative marginal log likelihood with 4,096 random features either using Algorithm S5 or using matrix decompositions, using a preconditioner rank of 512, a convergence threshold of 1e-5 and the "srht_2" preconditioner construction algorithm in all cases. We use either 40 probe vectors or 25 probe vectors; the distribution of the absolute percent error is plotted in Figure S4 below. 40 probe vectors does not provide a substantial improvement on 25 probe vectors. 25 probe vectors is sufficient in this experiment to achieve high accuracy.

**Figure S4.** Distribution of absolute percent error for marginal likelihood estimation across 112 evaluations using either 25 or 40 probe vectors.

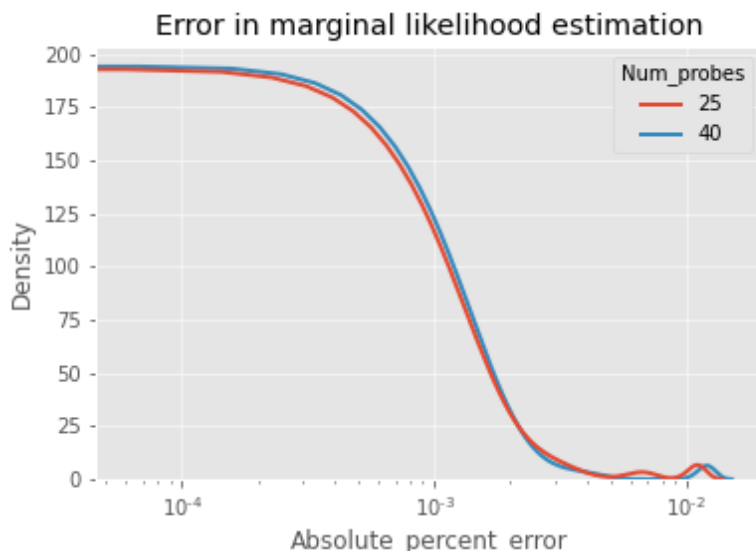

To further assess the accuracy of this procedure, we use approximate marginal likelihood in combination with Bayesian optimization to tune hyperparameters for two of the tutorials / example experiments in the user documentation (available at https://xgpr.readthedocs.io/en/latest/ ). The first such experiment involves the UCI proteins dataset, the second involves the QM9 dataset with one-hot encoded inputs. In both cases, the tuning procedure improves on the best validation set performance achieved by tuning with a smaller number of random features with a matrix decomposition-based approach. We further use the same approach to tune hyperparameters for the tabular data and the small molecule experiments on the QM9 dataset described in the main text and are able as described in the main text to achieve highly competitive performance. These results are consistent with the experiment from Figure S4 above and suggest that given an adequate preconditioner rank, a sufficiently small convergence threshold, and a sufficient number of probe vectors, for the datasets we consider here, Algorithm S5 is able to estimate the negative marginal log likelihood with high accuracy.

**S6. Improved strategies for hyperparameter tuning**

In section S5 above we showed how to use preconditioned conjugate gradients to calculate the marginal likelihood in an efficient way if the number of random features is large. Often however for noisy data we only need to use 1000 - 3000 random features to tune hyperparameters and get a result that provides sufficient performance. If we do indeed decide to use a smaller number of random features for tuning than for fitting, we can introduce a fast and highly efficient strategy. Recall that the marginal likelihood in the random features approximation is given by:

$$-\log p(y \mid X) = \frac{-1}{2\lambda^2} y^T Z (Z^T Z + \lambda^2 I)^{-1} Z^T y + \frac{1}{2\lambda^2} y^T y + (n-m)\log(\lambda) + \frac{1}{2}\log|Z^T Z + \lambda^2 I| + \frac{n}{2}\log(2\pi)$$

$Z^T Z$ is Hermitian so that as long as it is full-rank, we can generate an eigendecomposition $P\Lambda P^T$ such that P is orthonormal, i.e. $PP^T = I$. We then rearrange the above expression:

$$-\log p(y \mid X) = \frac{-1}{2\lambda^2} y^T Z P (\Lambda + \lambda^2 I)^{-1} P^T Z^T y + \frac{1}{2\lambda^2} y^T y + (n-m)\log(\lambda) + \frac{1}{2}\log|\Lambda + \lambda^2 I| + \frac{n}{2}\log(2\pi)$$

For three and four hyperparameter kernels, we can make use of this simplification in a variety of ways, for example Algorithm S3 and Algorithm S4.

## Algorithm S3  Efficient grid search

**Input:** Dataset X with N datapoints stored in chunks $x_1 \ldots x_n$ and $y_1 \ldots y_n$ on disk, initialized kernel with two, three or four hyperparameters ($\lambda$, $\beta$, $\sigma_1$, $if\ applicable\ \sigma_2$) for $M$ random features, selected values for hyperparameters (if applicable) $\sigma_1$ and $\sigma_2$

1. Initialize $M \times M$ array $W$ and $M \times 1$ array $S$, set to zero
2. $a = 0$
3. For j in 1…N:
    a. Load $x_j$, $y_j$ and use Algorithm S1 to form $Q_j = Z_j / \beta$ using $\delta_i$
    b. $W\mathrel{+}= Q^T Q$;  $S \mathrel{+}= Q^T y_i$
    c. $a \mathrel{+}= y_j^T y_j$
4. $U, \Lambda = eigh(W + \delta I)$     add small constant delta for numerical stability
5. $\Lambda \mathrel{-}= \delta$
6. $S = U^T S$
7. Define loss_fn($\beta$, $\lambda$):
    a. Return $\frac{-1}{2\lambda^2} S (\Lambda + \lambda^2 I)^{-1} S + \frac{1}{2\lambda^2} y^T y + (n - m)log(\lambda) +$
    $\frac{1}{2}log|\Lambda + \lambda^2 I| + \frac{n}{2}log(2\pi)$
8. $\lambda_{best}, \beta_{best} = argmin_{\lambda,\beta} loss$
9. $Score = loss(\lambda_{best}, \beta_{best})$

Return $Score$, $\lambda_{best}$, $\beta_{best}$

---

**Algorithm S4** Bayesian hyperparameter optimization (Thompson sampling)

---

**Input:** Dataset X with N datapoints stored in chunks $x_1 ... x_n$ and $y_1 ... y_n$ on disk, initialized kernel with three or four hyperparameters ($\lambda$, $\beta$, $\sigma_1$, $if\ applicable\ \sigma_2$) for *M* random features, *maxiter* maximum iterations, *n* candidates per iteration, convergence criteria *tol*

1. Evaluate score, $\lambda_{best}$, $\beta_{best}$ for 5 - 10 randomly selected kernel-specific hyperparameters $\omega_1 ... \omega_{10}$ using Algorithm S3
2. Let $\Omega$ be the set of all kernel-specific hyperparameter combinations evaluated thus far; let $\omega_{best}$ be the best evaluated so far
3. For niter in 10…maxiter:
    a. Fit exact Gaussian process *S* with Matern kernel, $\nu = 5/2$
    b. Draw *n* kernel-specific hyperparameter combinations $\omega_1 ... \omega_n$ from $Uniform(\sigma_{max}, \sigma_{min})$
    c. Draw *m* samples from *S* at each of $\omega_1 ... \omega_n$; select $\omega_{new} = argmin_\omega (m)$
    d. If $||\omega_{new} - \omega_{best}||_2 < tol$:
        i. break
    e. Evaluate the loss function from Algorithm S3 at $\omega_{new}$
    f. If $loss_{new} < loss_{best}$:
        i. $\omega_{new} = \omega_{best}$
4. Return $\omega_{best}$

---

Note that we use Thompson sampling in S4, but of course other acquisition functions (e.g. expected improvement) can easily be substituted as well. We can also use a simple grid search across the kernel-specific hyperparameter, which is also implemented in xGPR. We can also modify the above algorithm by calculating the eigendecomposition of $Z^T Z$ by taking the singular value decomposition of $Z$, which is of course the most stable approach.

The polynomial kernel has only two hyperparameters (the lambda or noise and beta or amplitude hyperparameters shared with all other kernels). Using the strategy above, these can be tuned using a *single pass over the data*, so that hyperparameter tuning with the polynomial kernel is exceptionally fast.

## S7. Datasets used for experiments described in the main text

All datasets used in the paper are publicly available. Details of each dataset are described in the following table.

**Table S3. Details of datasets used for experiments in the main text**

| Dataset name | Data type | Number of datapoints | Task | Link | Citation |
|---|---|---|---|---|---|
| UCI SONG | Tabular data | 515,345 | Predict the year in which a song was released based on features of the song. | https://archive.ics.uci.edu/ml/datasets/yearpredictionmsd | Bertin-Mahieux et al.[1] |
| UCI Proteins | Tabular data | 45,730 | Predict the RMSD of a residue based on physicochemical properties. | https://archive.ics.uci.edu/ml/datasets/Physicochemical+Properties+of+Protein+Tertiary+Structure | Qian and Sejnowski 1988.[2] |
| UCI Kin40K | Tabular data | 40,000 | Predict the forward motion of a robot arm using 8 inputs | https://www.cs.utoronto.ca/~delve/data/kin/desc.html | Corke 1996.[3] |
| AAV | Sequence data | 201,426 | Predict fitness properties of mutant sequences of the AAV capsid protein | https://benchmark.protein.properties/landscapes | Dallago et al.[4] |
| GB1 | Sequence data | 8,733 | Predict fitness properties of mutants of the binding domain of protein G from Streptococcus | https://benchmark.protein.properties/landscapes | Dallago et al.[4] |
| Thermostability | Sequence data | 27,951 | Predict thermostability of a diverse array of proteins from different families | https://benchmark.protein.properties/landscapes | Dallago et al.[4] |
| Fluorescence | Sequence data | 54,025 | Predict fluorescence of mutants of GFP | https://github.com/songlab-cal/tape | Rao et al.[5] |
| Stability | Sequence data | 68,965 | Predict stability in a protease degradation assay of mutant protein sequences. (This is completely unrelated to the thermostability dataset of Dallago et al and while diverse is significantly less so.) | https://github.com/songlab-cal/tape | Rao et al.[5] |
| QM9 | Graph / small molecule data | 134,000 | Predict DFT-calculated properties of small molecules given the DFT-optimized gas-phase geometry | https://figshare.com/collections/Quantum_chemistry_structures_and_properties_of_134_kilo_molecules/978904 | Ramakrishnan et al.[6] |

## S8. Comparison of xGPR with stochastic variational inference and exact Gaussian processes

The stochastic variational approximation to Gaussian process models and variants have appeared frequently in the literature[7,8]. From a theoretical standpoint it has many attractive properties, but how does it perform in practice? In the table below, we compare the performance of SVGP as implemented in the GPyTorch library with the random feature approximation as implemented in xGPR; for sufficiently small datasets, we compare also with an exact Gaussian process as implemented in GPyTorch. (We tried to train an exact Gaussian process for larger datasets but met with an out of memory error for the UCI proteins dataset with about 8,000 datapoints, so that for even larger datasets it is clearly impractical). We train stochastic variational inference in GPyTorch for 40 epochs in all cases, and we use the RBF kernel in all cases, since we are not aware of an SVGP-compatible convolution kernel for sequence data sufficiently efficient that we can use it on a > 100,000 sequence dataset in a short period of time, nor are any such kernels implemented in GPyTorch. Training for 20 epochs yielded worse performance in every case.

For xGPR, we tune hyperparameters using 3,000 random features using the "fast" strategy in Algorithm S3 from section S6 above. We then fit using either 16,384 or 32,768 random features for tabular data and 8,192 or 16,384 random features for sequence data (we use a smaller number of random features for sequence data to correspond to the number of features used for comparison with CNN deep learning models on sequence data in the main text). Note that several of the datasets used as benchmarks here are protein sequence datasets also used as benchmarks for comparisons with CNNs in the main text. In the main text, we use sequence kernels that do not require the input to be aligned. Here, by contrast, we are using an RBF kernel (since no sequence kernels are available in GPyTorch or scikit-learn). Thus, in this comparison, for the protein sequence datasets, we convert each into a multiple sequence alignment and one-hot encode it before using it as input to an RBF kernel, in contrast to how the sequence datasets are modeled in the main text.

Table S1 clearly shows that xGPR outperforms SVGP on 8 out of 9 datasets, often by large margins. In the case where SVGP is competitive (Song UCI), they achieve essentially equivalent performance, and SVGP takes eighty times longer to train. xGPR generally performs better using the same kernel and at greatly reduced cost; moreover, performance for xGPR can often be further improved by either using more random features for fitting, or by fine-tuning hyperparameters with a larger number of random features. On the Kin40K dataset, for example, "fine-tuning" the hyperparameters using the marginal likelihood approximation strategy from section S5 above requires an additional 10 minutes and reduces the mean absolute error by an additional 5-8% (depending on the number of random features used to fit). Improving the performance achieved by SVGP, by contrast, would be hard, since if the number of inducing points for SVGP were increased, training time on the larger datasets would become unacceptable.

It is possible that performance for SVGP could be improved by modifying the Adam optimizer settings for stochastic gradient descent, e.g. by dataset-specific tuning of the hyperparameters of the Adam algorithm, but this is in one sense merely a further drawback for SVGP: xGPR does not require extensive experimentation with optimizer settings to achieve good outcomes. Given these results, we did not implement stochastic variational inference for our library, preferring our approach instead.

**Table S4. Performance of SVGP (as implemented in the GPyTorch library), random features (as implemented in xGPR) and exact GPs (as implemented in scikit-learn) for various benchmark datasets**

| Dataset | Split | # training / test datapoints | Metric | GPyTorch SVGP RBF training time (min), 500 inducing points / 3,000 inducing points | GPyTorch score, 500 / 3000 inducing points | xGPR, RBF kernel, training time (min), low / high fitting RFFs | xGPR, RBF kernel, score, low / high fitting RFFs | Exact Gaussian process, RBF kernel, score (if applicable) |
|---|---|---|---|---|---|---|---|---|
| Kin40K (UCI) | NA | 53,614 / 12,581 | MAE (lower is better) | 0.5 / 27 | 0.17 / 0.10 | 1.5 / 1.8 | **0.099 / 0.087** | |
| Song (UCI) | Mixed | 412,276 / 103,069 | MAE (lower is better) | 6.4 / 348 | 0.57 / 0.57 | 6.0 / 6.9 | 0.57 / 0.57 | |
| Protein (UCI) | Human only | 36,584 / 9,146 | MAE (lower is better) | 0.5 / 31 | 0.56 / 0.50 | 0.69 / 0.78 | **0.48 / 0.47** | |
| Alpha-adenovirus (AAV) | Designed vs mutant | 201,426 / 82,583 | Spearman's r (larger is better) | 3.7 / 158 | 0.58 / 0.63 | 3.4 / 3.8 | **0.73 / 0.74** | |
| Alpha adenovirus (AAV) | Mutant vs designed | 82,583 / 201,426 | Spearmans' r (larger is better) | 1.5 / 65 | 0.67 / 0.65 | 2.9 / 3.1 | **0.76 / 0.78** | |
| Alpha adenovirus (AAV) | 7 mutations vs many | 70,002 / 12,581 | Spearmant's r (larger is better) | 1.2 / 54 | 0.58 / 0.62 | 1.9 / 2.3 | **0.67 / 0.68** | |
| Fluorescence | NA | 21,446 / 27,217 | Spearmant's r (larger is better) | < 1 / 21 | 0.66 / 0.67 | 1.7 / 1.8 | **0.68 / 0.68** | |
| GB1 | 3 vs rest | 2,968 / 5,765 | Spearmant's r (larger is better) | No data / <1 | No data / 0.82 | < 1 | 0.82 / 0.83 | **0.84** |
| GB1 | 2 vs rest | 427 / 8,306 | Spearmant's r (larger is better) | < 1 / No data | 0.48 / No data | < 1 | 0.62 / 0.62 | **0.63** |

### S9. Lemma 1.1

*Lemma 1.1.* Let $K: \mathbb{R}^D \times \mathbb{R}^D \rightarrow \mathbb{R}$ be a positive definite kernel on $\mathbb{R}^D$, and let $f: X \rightarrow \mathbb{R}^D$ be any mapping from $X$ to $\mathbb{R}^D$, where $X$ is some non-empty set. Then $K(f(u), f(w))$ is a positive definite kernel on $X$ for $u, w \in X$.

Proof: Let $f: X \rightarrow \mathbb{R}^D$ be any mapping from $X$ to $\mathbb{R}^D$, where $X$ is a non-empty set. Let $K: \mathbb{R}^D \times \mathbb{R}^D \rightarrow \mathbb{R}$ be a positive definite kernel on $\mathbb{R}^D$, i.e.:

$$\sum_i^N \sum_j^N c_i c_j K(x_i, x_j) \geq 0 \qquad (1.1a)$$

for any choice of $x_1, x_2, x_3 ... x_n \in \mathbb{R}^D$ and for any choice of $c_1, c_2, c_3 ... c_n \in \mathbb{R}$. Since this is true for any arbitrary selection of $x_1, x_2, x_3 ... x_n \in \mathbb{R}^D$, and since $f$ maps its input into $\mathbb{R}^D$, it is also true that

$$\sum_i^N \sum_j^N c_i c_j K(f(x_i), f(x_j)) \geq 0$$

for any choice of $x_1, x_2, x_3 ... x_n \in X$.

In other words – this is merely another way to say that any feature engineering process or function which constructs a representation in $\mathbb{R}^D$ can be "plugged into" a kernel which is positive-definite on $\mathbb{R}^D$ and retain the positive definite property. We can use this to construct "plug-in" kernels, e.g. the FastConv1d kernel in the text.

**S10. Details for QM9 molecular modeling**
The full QM9 dataset consists of approximately 133,000 molecules. To be consistent with other authors, we first remove those molecules for which Ramakrishnan et al. indicate that the geometry-optimized structure failed consistency checks. We then split the remaining data randomly into a training set of 110,000, a validation set of 10,000, and a test set of 10,831 using random seed 123.

For the SOAP descriptors as implemented in the dscribe library, there are several hyperparameters we can tune:

1) The standard deviation of the Gaussians used to represent the atoms in the neighborhood of the central atom (sigma);
2) *Rcut*, or the radius at which an atom is considered "outside the neighborhood" and no longer included in the representation;
3) Weighting, i.e. reduced weighting for atoms further from the central atom (several different schemes provided).

There are also the number of radial basis functions, *n_max*, and the number of spherical harmonics, *l_max*. These improve the accuracy of the representation if increased, but also dramatically increase the size of the descriptor vector. We settled on 12 for n_max and 9 for l_max, the same values used by Willatt et al.[22], since we did not want to use a larger descriptor vector size for this study. Even this configuration however generates roughly 18,000 features per atom.

We started from the weighting scheme and weights identified as optimal by Willatt et al.[22], who used power weighting in which an atom within the distance *rcut* is weighted based on its distance using the expression:

$$\frac{1}{1+(\frac{r}{r_0})^m}$$

For all of the SOAP feature hyperparameters we used the values selected by Willatt et al. based on their grid search with two exceptions: we optimized *rcut*, or the cutoff distance, and *r0* from the weighting scheme shown above, using the values from Willatt et al. as a starting point. We tuned *r0* and *rcut* by evaluating the marginal likelihood on a randomly selected subset of the training data consisting of 25,000 datapoints. In initial experiments, we found that we could achieve almost the same accuracy by generating features for heavy atoms (i.e. non-hydrogen) only and omitting the hydrogens. This is advantageous since otherwise the amount of disk space required to store the training set data is quite large. After tuning, we adopted the following settings for dscribe:

N_max: 12
L_max: 9
Sigma: 0.25
R_cut: 3.25
Weighting: Power
R0: 1.5
M: 9

*M* and *r0* determine how rapidly the weighting falls off from the central atom. We generate SOAP features for each atom individually; since each molecule in this dataset has up to 9 heavy atoms, each molecule has up to 9 descriptor vectors corresponding to these atoms. These can be used as inputs to either the graph convolution kernel or graph polynomial kernel described in the main text. We divide the SOAP descriptors for each atom by the norm of the descriptor vector so that each descriptor vector is unit-norm; this ensures that if the GraphPoly kernel is used, it will have the same interpretation as the original SOAP kernel.

To tune the kernel hyperparameters, we use the workflow described under S5 and S6 above. We begin by finding a starting point using matrix decomposition based marginal likelihood with a small number of random features, then optimize with a larger number of random features (16,384) using Bayesian optimization. This procedure took 4 hours on a single A6000 GPU. All hyperparameter tuning was performed using internal energy at 298K as a target. The same hyperparameters that work well for internal energy at 298K also work well for the other objectives as illustrated. We then fit the final model using either 16,384 random features, 32,768 random features or 66,536 random features. The results appear in Table 4 of the main text.

As expected, we can "buy" small improvements in performance by using a larger number of random features, but even 16,384 random features is sufficient to achieve a mean absolute error 5x smaller than chemical accuracy. Consequently, in practice using larger numbers of random features has limited usefulness and it would be perfectly fine to use 16,384 random features for this problem. Moreover, the most important target for further improvement is likely the SOAP features themselves. The large number of SOAP features generated per atom greatly increases computational expense, since many features must be generated, saved and loaded from disk while processing each minibatch of data when iterating across the dataset. Furthermore, it constrains our ability to improve the accuracy of the model by increasing the number of basis functions. Multiple authors have suggested more compact representations that achieve similar or better accuracy. A more compact representation would enable us to increase the number of basis functions, thereby improving the accuracy of the model, while decreasing the length of the feature vector, thereby reducing its computational expense. We regard this as an important area for further work.

It is interesting to note the advantages provided by our fast Hadamard transform based Conv1d kernel (FHTConv1d) here as well. If we were to use vanilla random features instead, to generate 66,536 random features with 18,000 input features, we would require a matrix with roughly 1.2 billion floats! The FHTConv-1d kernel, by contrast, needs to store roughly 288,000 8-bit integers and 72,000 floats – a dramatic reduction in memory footprint.

**S11. Clustering and kernel PCA using representations generated by xGPR**

For this example, we use the QM9 dataset. Out of the approximately 130,000 molecules in the dataset, we randomly select 20,000 as a test set. We encode each atom and its neighbors using one-hot encoded, out to the 15th most distant neighbor (an arbitrary cutoff). Each one-hot encoded neighbor is weighted by 1 / distance**6; the 6th power used here corresponds to the weighting used in the London matrix sometimes used as a representation for machine learning on small molecules. We then follow the standard workflow by tuning hyperparameters first using a smaller number of random features (2048) with a matrix-decomposition based procedure, then "fine-tune" using a larger number of random features (8192) with an approximate marginal likelihood procedure. The code used to encode the molecules and tune the hyperparameters is provided in the small molecule tutorial in the xGPR documentation, available online at https://xgpr.readthedocs.io/en/latest/ . As illustrated, depending on the number of random features employed for fitting, this model can achieve a mean absolute error < 1.1 kcal/mol.

Next, we use the kernel_xpca tool provided in the visualization_toolkit module of xGPR version 0.0.2.1 to retrieve the top 500 principal components. We use the top 2 principal components to generate the kernel PCA plot below in Figure S5, for the test set data only.

**Figure S5.** Kernel PCA on the QM9 test set, using the GraphConv-1d kernel

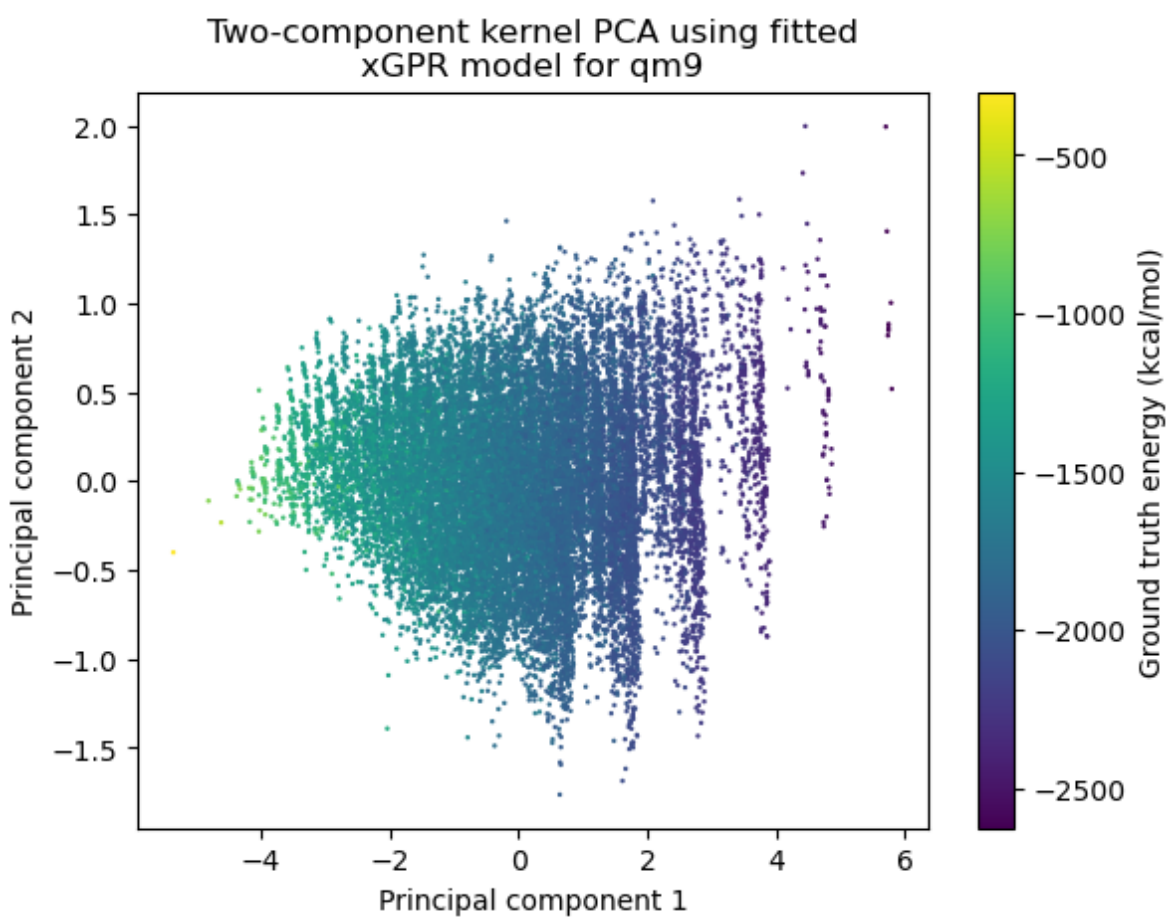

As illustrated, the top principal component correlates nicely with the ground-truth label we want to predict (energy at 298K). This will not always be the case, because a great deal of information is obviously discarded by using a 2d representation in place of the 16,384 dimensional representation generated by xGPR. To further explore the data, then, we perform k-means clustering using the scikit-learn library with the top 500 principal components as input, using the sum squared error to assess the performance of a given number of clusters. This operation approximates performing kernel k-means clustering on the original training data using a GraphConv-1d kernel.

The elbow plot appears in Figure S6a; this result suggests 5 clusters may be appropriate, and thus we fit the final k-means model using 5 clusters. The distribution of energy at 298 K in each resulting cluster appears in Figure S6b. In many drug discovery scenarios this ability to cluster the data using the same model used to fit it might in itself be useful; we further discuss how it can be used for efficient retrieval of similar molecules from the training set in the main text.

**Figure S6.** Kernel k-means clustering of the QM9 training set, using the GraphRBF kernel

A

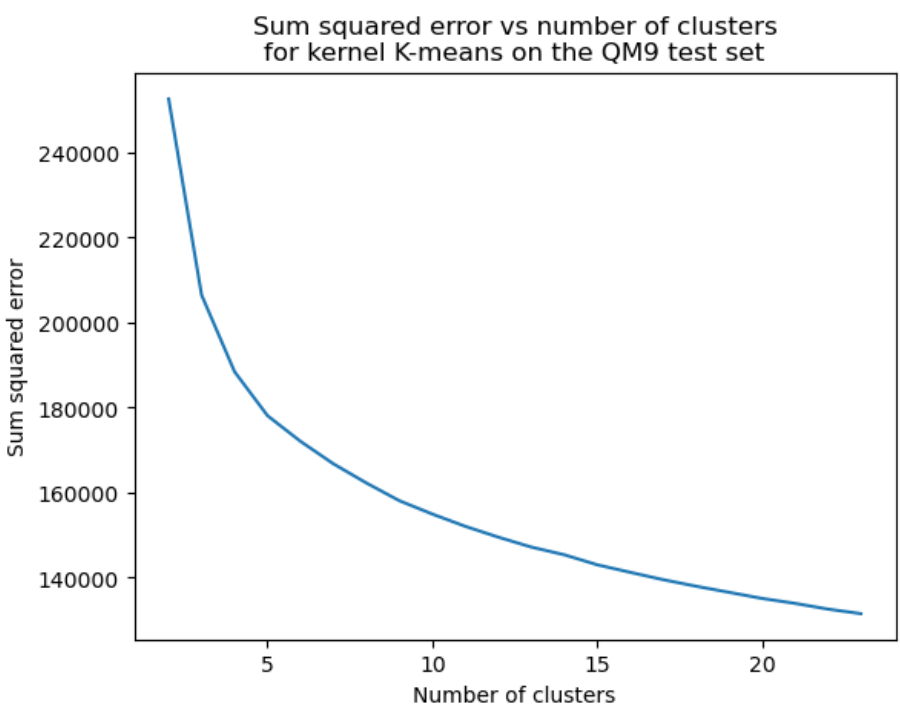

B

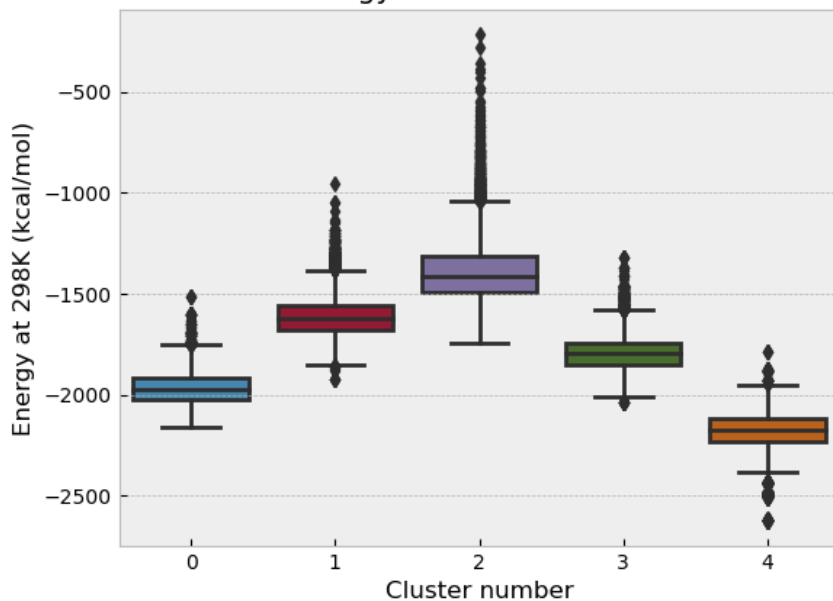

## S12. Molecules from the test set and the most similar molecules from the training set.

Five random molecules from the QM9 test set are depicted in Figures S7, S8, S9, S10 and S11 below, together with the 50 most similar molecules from the training set, where similarity is quantified using the (approximated) GraphRBF kernel function with molecules represented using one-hot encoding as described in the main text. Note that this representation of molecules is relatively uninformative, so it is to some extent surprising that the model performs as well as it does (MAE of about 1 kcal/mol) and can retrieve molecules that in many cases (as illustrated below) do show many points of similarity.

**Figure S7. Test query molecule 1, together with the most similar training molecules**

**Figure S8. Test query molecule 2, together with the most similar training molecules**

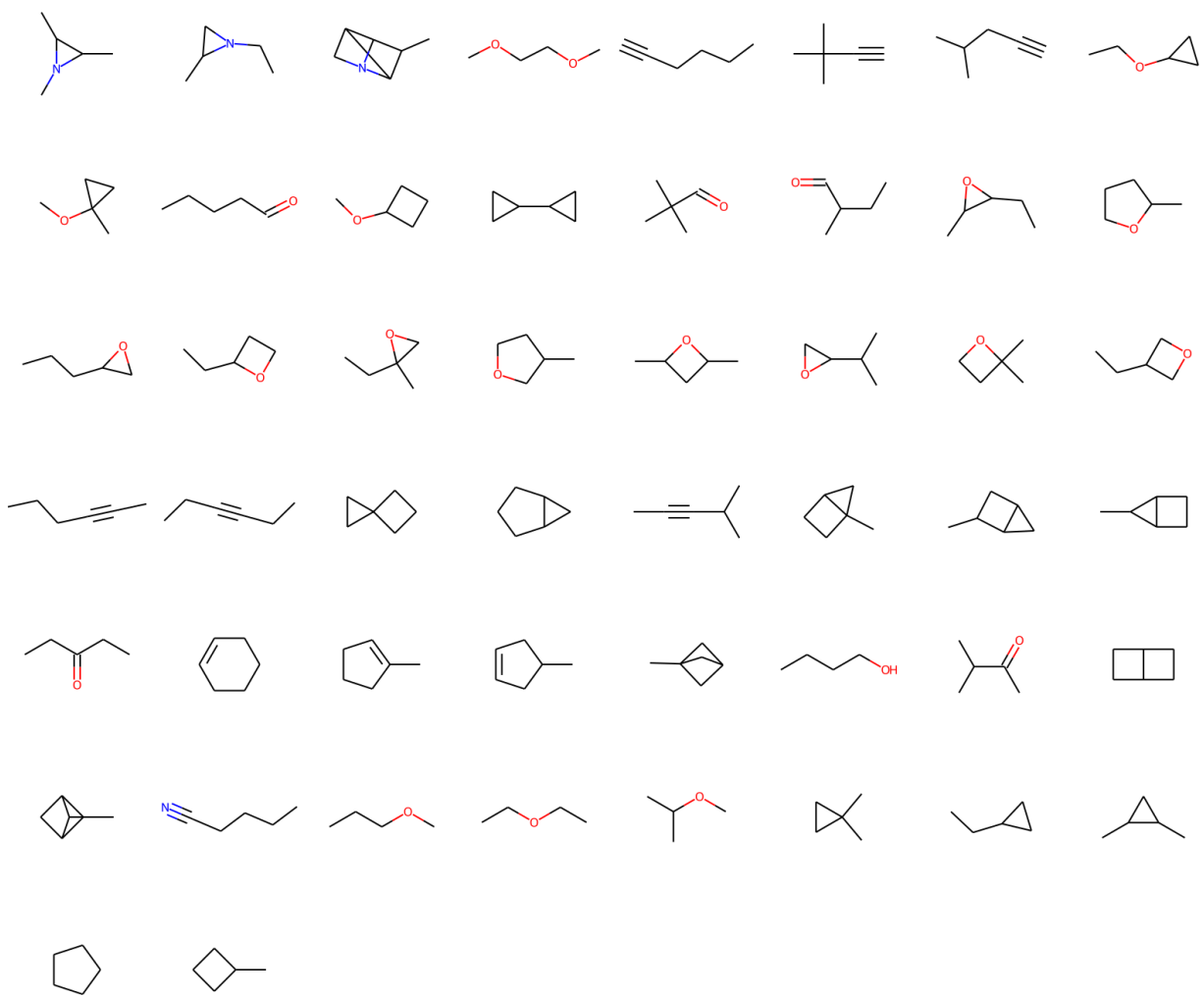

**Figure S9. Test query molecule 3, together with the most similar training molecules**

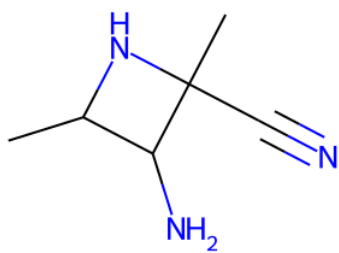
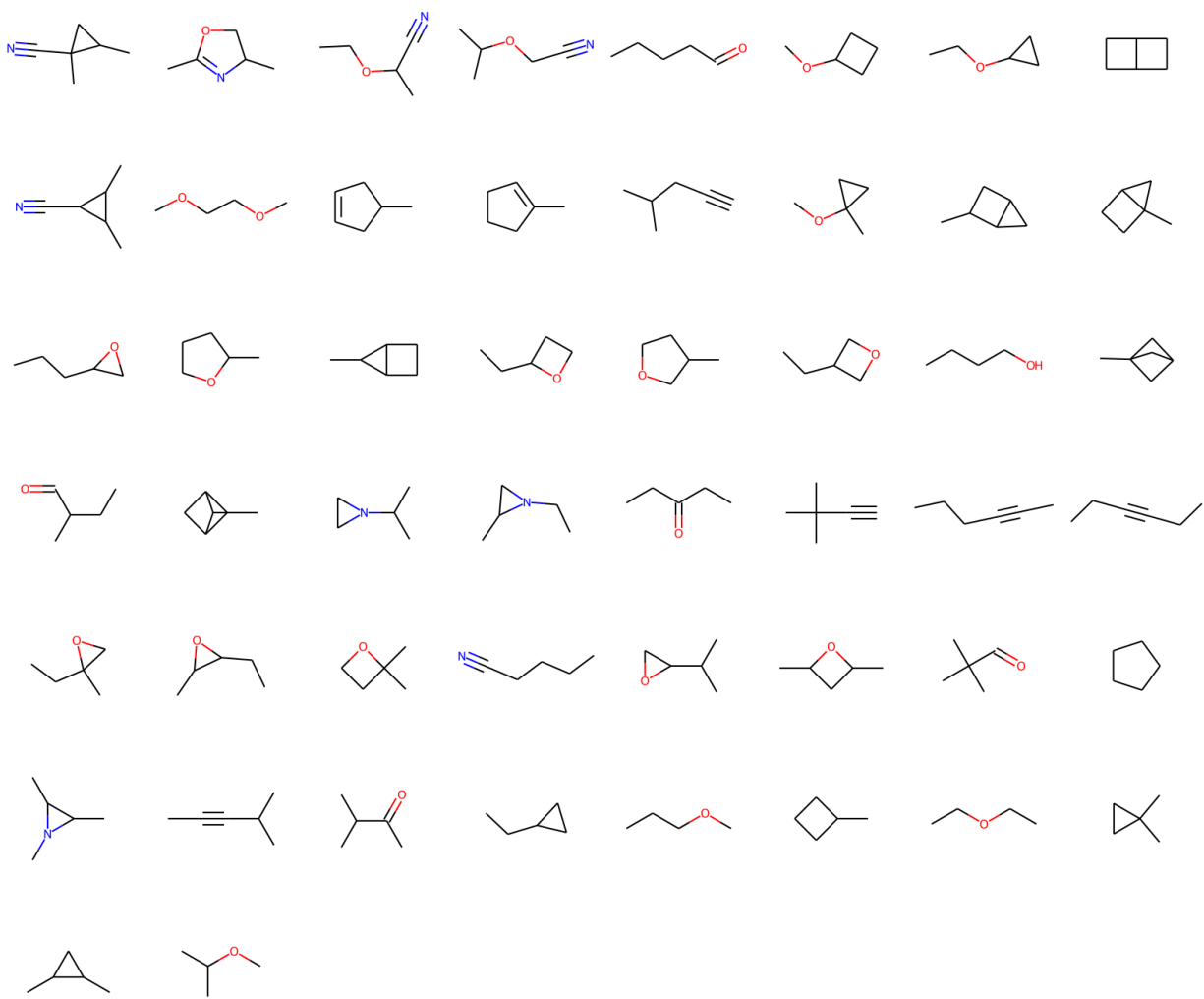

**Figure S10. Test query molecule 4, together with the most similar training molecules**

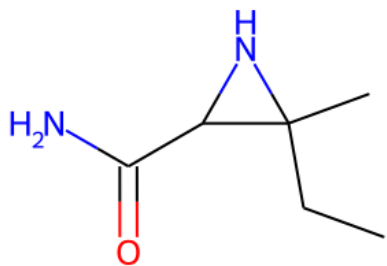

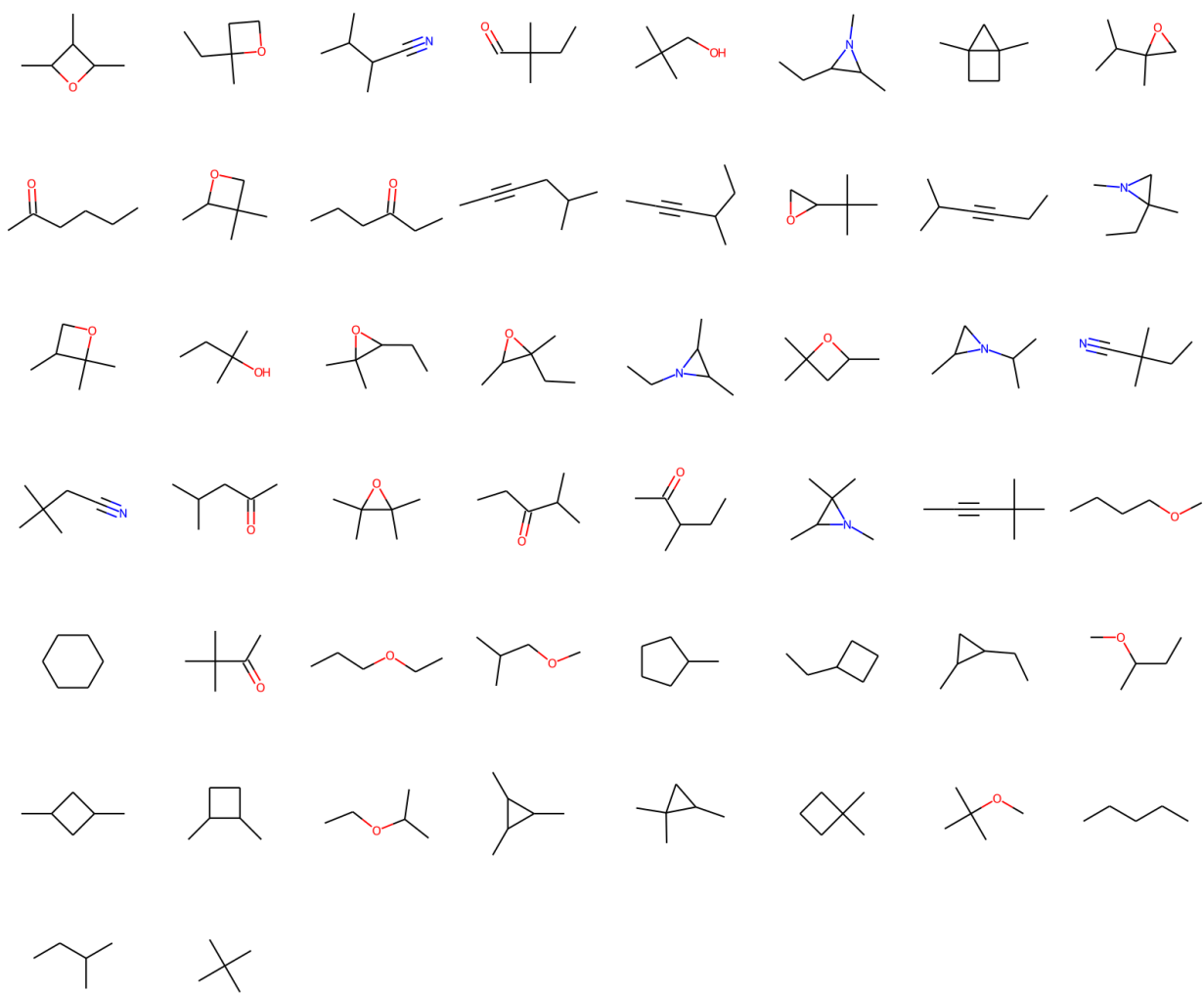

**Figure S11. Test query molecule 5, together with the most similar training molecules**

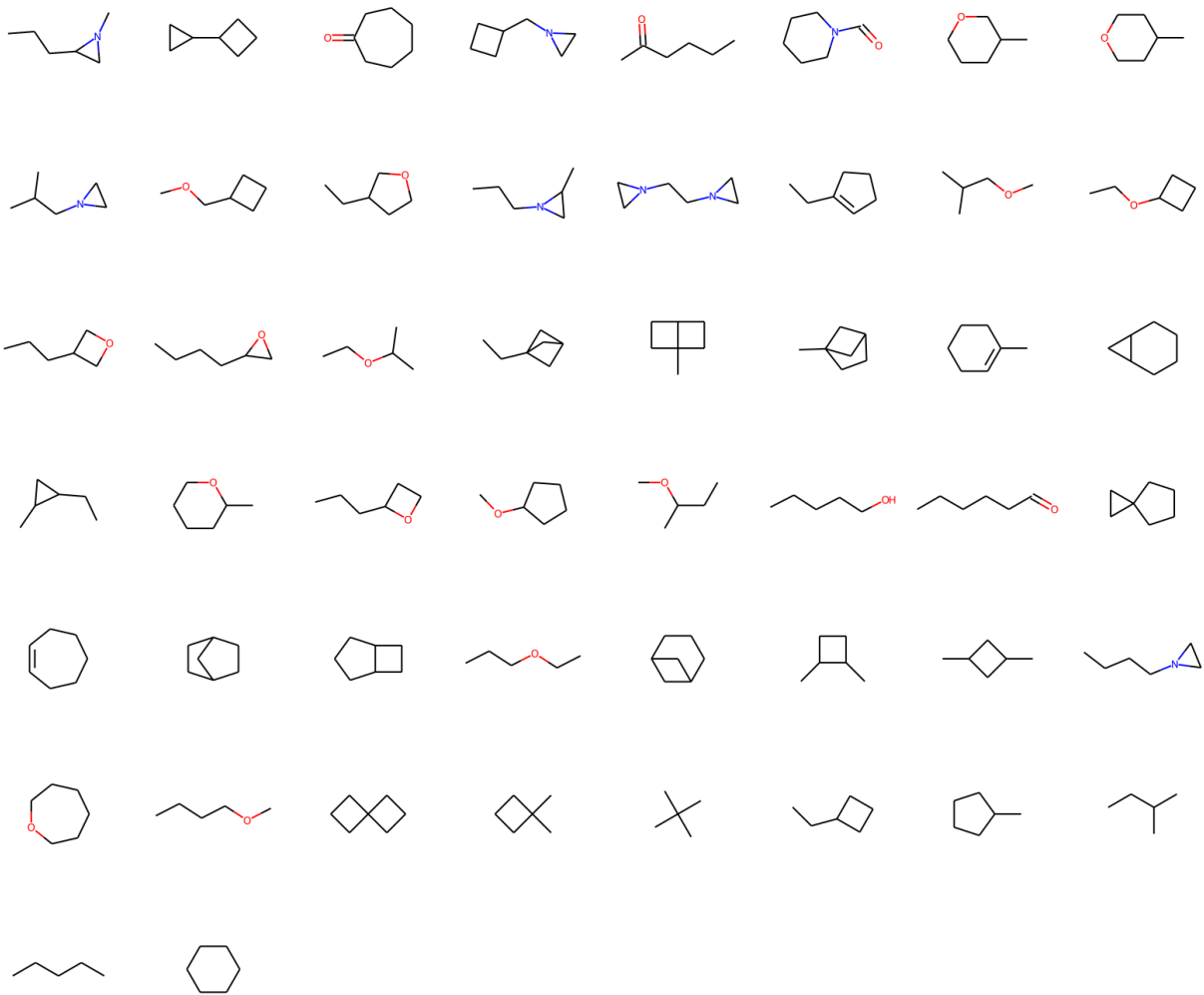